\def\year{2022}\relax
\documentclass[letterpaper]{article} % DO NOT CHANGE THIS
\usepackage{aaai22}  % DO NOT CHANGE THIS
\usepackage{times}  % DO NOT CHANGE THIS
\usepackage{helvet}  % DO NOT CHANGE THIS
\usepackage{courier}  % DO NOT CHANGE THIS
\usepackage[hyphens]{url}  % DO NOT CHANGE THIS
\usepackage{graphicx} % DO NOT CHANGE THIS
\usepackage{comment}
\usepackage{natbib}  % DO NOT CHANGE THIS AND DO NOT ADD ANY OPTIONS TO IT
\usepackage{caption} % DO NOT CHANGE THIS AND DO NOT ADD ANY OPTIONS TO IT
\DeclareCaptionStyle{ruled}{labelfont=normalfont,labelsep=colon,strut=off} % DO NOT CHANGE THIS
\usepackage{algorithm}
\usepackage{algorithmic}
\usepackage{booktabs}
\usepackage{subfigure}
\usepackage{multirow}
\usepackage{epstopdf}
\usepackage{amsmath}
\usepackage{adjustbox}
\usepackage{newfloat}
\usepackage{listings}
\floatstyle{ruled}
\newfloat{listing}{tb}{lst}{}
\floatname{listing}{Listing}
\title{A Comparative Study on Robust Graph Neural Networks to Structural Noises}
\author{
    Zeyu Zhang,
    Yulong Pei
}
\affiliations{
    Eindhoven University of Technology\\
    Eindhoven, the Netherlands\\
    z.zhang3@student.tue.nl,y.pei.1@tue.nl
}
\usepackage{bibentry}
\begin{document}

\maketitle

\begin{abstract}
Graph neural networks (GNNs) learn node representations by passing and aggregating messages between neighboring nodes. GNNs have been applied successfully in several application domains and achieved promising performance. However, GNNs could be vulnerable to structural noise because of the message passing mechanism where noise may be propagated through the entire graph. Although a series of robust GNNs have been proposed, they are evaluated with different structural noises, and it lacks a systematic comparison with consistent settings. In this work, we conduct a comprehensive and systematical comparative study on different types of robust GNNs under consistent structural noise settings. From the noise aspect, we design three different levels of structural noises, i.e., local, community, and global noises. From the model aspect, we select some representative models from sample-based, revision-based, and construction-based robust GNNs. Based on the empirical results, we provide some practical suggestions for robust GNNs selection.
\end{abstract}

\section{Introduction}
Graph-structured data is ubiquitous in plenty of real-world applications for its effectiveness in representing complex interactions. As a powerful tool for analyzing graph-structured data, Graph Neural Networks (GNNs) have been widely studied in the past years, which have led to state-of-art results on several tasks, including node inference, link prediction, information retrieval, etc.

Nevertheless, traditional GNNs are usually problematic when deployed to sensitive areas such as finance and healthcare because of robustness reasons. Like most neural network-based learning approaches, the noise that appears in data will lead to a significant performance degeneration for GNNs, especially when encountering structural noise. Most GNN models utilize the message passing mechanism as their fundamental, as each node will iteratively update its embedding by aggregating with its neighbors' embeddings. However, this procedure can have cascading effects when the graph data is noised, since the error may be propagated through the entire graph. Several studies have been conducted to address the robustness issue of GNNs. There are three major types of approaches to tackle this problem, which can be categorized into sample-based, revision-based, and construction-based methods shown in Table~\ref{tb:taxonomy}. 

Adapting the sampling method in GNNs is usually feasible to enhance the robustness. Considering the characteristics of graph data, there are two main choices: sampling edges and sampling nodes. DropEdge \cite{rong2019dropedge}, as a representative of edge-sampling method, randomly removes a certain number of edges from the input graph at each training epoch. DropCONN \cite{chen2020enhancing} inherits this method but utilizes a biased-sampling scheme to drop graph connections such that subgraphs are constructed for training and inference. As a node-sampling method, GraphSAGE \cite{hamilton2017inductive} learns a function that generates embedding by sampling and aggregating features from a node’s local neighbors. FastGCN \cite{chen2018fastgcn} interprets graph convolutions as integral transforms of embedding functions, which enables using Monte Carlo approach to estimate the integrals and leads to a batched training/inference scheme.

Revision-based methods aim to produce a denoised graph structure from the noised one. Applying metric learning \cite{zhu2021deep} is a typical way to achieve this goal. For example, GRCN \cite{yu2020graph} introduces a GCN-based graph revision module for predicting missing edges and revising edge weights w.r.t. downstream tasks via joint optimization. RGCN \cite{zhu2019robust} adopts Gaussian distributions as the hidden representations of nodes in each convolutional layer to absorb the effects of adversarial changes. Another possible way is to use direct optimization. A representative tool is Pro-GNN \cite{jin2020graph} that cleans the perturbed graph guided by several graph’s intrinsic properties: low rank, sparsity, and feature smoothness. This framework jointly learns the graph structure and a robust GNN.

Construction-based approaches intend to learn a new graph representation that captures most interactions between nodes. It has two schemes: bottom-to-up and up-to-bottom. LDS~\cite{franceschi2019learning}, as a bottom-to-up method, jointly learns the graph structure and the parameters of graph convolution networks by approximately solving a bilevel program that learns a discrete probability distribution on the edges of the graph. On the other hand, PTDNet \cite{zheng2020robust}, as a up-to-bottom method, aims to remove unnecessary edges for graph compression while keeping almost all information of the input graph.

Although many models have been proposed to improve the robustness of GNNs recently, they are usually designed for specific noise settings. This inconsistency often leads to distinct even conflict results when comparing the performance in different studies. Considering that the structural noises could stem from different levels including local and global, the lack of systematical comparison on these approaches limits the selection of suitable robust GNNs in practice. Although important, to the best of our knowledge, there are not studies that systematically compare the performance of different robust GNN models under the same structural noise settings. The only related work is~\cite{fox2019robust}, but it only tests one method, i.e.,~GIN~\cite{xu2018powerful}, on synthetic networks. In this paper, we conduct a more comprehensive and systematical comparison study on several different types of robust GNNs under different levels of structural noises.

To sum up, in this paper we have following major contributions: (1) we design three different levels of graph structural noises to test the robustness of GNNs, (2) we empirically evaluate three types of robust GNNs under consistent noise settings, and (3) we provide some practical suggestions for selecting suitable robust GNNs towards different levels of structural noises.

\section{Preliminaries}

\subsection{Graph Structure Mining}
Graphs are widely used in representing complex relationships among entities, for instance, social networks, biological networks, and traffic networks. As graphs are seemingly ubiquitous in different domains, it is of great theoretical and practical value to study graphs. In order to analyze graphs, one of the important tasks is to grasp the structures of the target graphs. Graph structures can be analyzed locally or globally from different perspectives.
\begin{figure}[htb]
\vskip -0.1in
  \centering
  \includegraphics[width=0.42\textwidth]{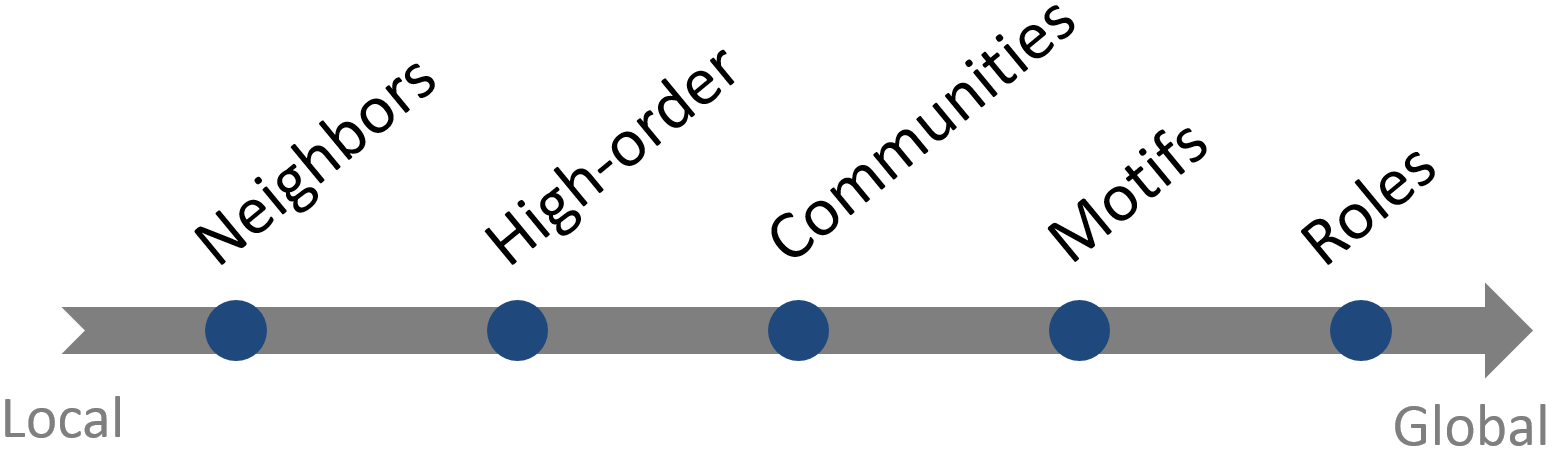}
  \caption{Different levels of graph structures.}
  \label{fig:struct}
  \vskip -0.1in
\end{figure}

Not strictly, from the local to global perspective, different levels of graph structures are ranged from neighbors, high-order connections, communities, motifs to roles shown in Figure~\ref{fig:struct}. In specific, the most straightforward local structure is reflected by the direct connections, i.e., neighborhood information. High-order structures include k-hop connections where $k\ge 2$. Community is a set of nodes that are densely connected internally. Motifs are recurrent and statistically significant subgraphs or patterns of a graph which can capture some globally shared structural information. For global structures, studies mainly focus on role discovery and analytics~\cite{rossi2014role}. Roles can represent node-level connectivity patterns, e.g., bridge, cliquey, isolated.

\subsection{Graph Neural Networks}
Graph neural networks (GNNs)~\cite{kipf2016semi,hamilton2017inductive,velivckovic2017graph,xu2018powerful} learn node representations by passing and aggregating messages between neighboring nodes. Following~\cite{xu2018powerful}, a $k$-layer GNN is formally defined as:
\begin{align}
\small
    a_v^{(k)}&=AGGREGATE^{(k)}\Big(\Big\{h_u^{(k)}:u\in N(v)\Big\}\Big)\\\nonumber
    h_u^{(k)}&=COMBINE^{()k)}\Big(h_v^{(k-1)},a_v^{k}\Big)
\end{align}
where $h_v^{(k)}$ is the feature vector of node $v$ at the $k$-th iteration/layer. Initially, we have $h_v^{(0)}=X_v$ and $N(v)$ is neighbors of node $v$.

However, since GNNs learn representations based on message passing via edges and aggregating from neighbors, graph structures with noisy and incomplete information may degenerate the performance of GNNs in learning representations and downstream tasks. In practice, noisy structures are often inevitable when collecting graph data. For instance, network delay may result in missing edges and spammers may create edges that should not exist. To overcome this challenge, some robust GNNs have been proposed and we categorize these GNNs into three types: sample-based, revision-based, and construction-based, shown in Table~\ref{tb:taxonomy}. The selected methods in our comparative study are highlighted in bold in this table.
\begin{table*}
\caption{A taxonomy of robust Graph Neural Networks.}
	\label{tb:taxonomy}
	%\vskip -0.1in
	\begin{center}
		\begin{small}
			\begin{sc}
			    \resizebox{\textwidth}{!}{
\begin{tabular}{l|l|l|l}
\hline
\multicolumn{2}{c}{type}                                                                                             & Method                         & Implementation \\ \hline
\multicolumn{1}{l|}{Sample-based}                        & \multicolumn{1}{l|}{Edge}                                 & \multicolumn{1}{l|}{\textbf{DropEdge}~\cite{rong2019dropedge}}  & \url{https://github.com/DropEdge/DropEdge}           \\ \cline{3-4} 
\multicolumn{1}{l|}{}                                    & \multicolumn{1}{l|}{}                                     & \multicolumn{1}{l|}{DropCONN~\cite{chen2020enhancing}}  & Code not available           \\ \cline{2-4} 
\multicolumn{1}{l|}{}                                    & \multicolumn{1}{l|}{\multirow{2}{*}{Node}}                & \multicolumn{1}{l|}{\textbf{GraphSAGE}~\cite{hamilton2017inductive}} &  \url{https://github.com/williamleif/GraphSAGE}        \\ \cline{3-4} 
\multicolumn{1}{l|}{}                                    & \multicolumn{1}{l|}{}                                     & \multicolumn{1}{l|}{\textbf{FastGCN}~\cite{chen2018fastgcn}}   &  \url{https://github.com/Gkunnan97/FastGCN_pytorch}   \\ \hline
\multicolumn{1}{l|}{\multirow{4}{*}{Revision-based}}     & \multicolumn{1}{l|}{\multirow{2}{*}{Metric Learning}}     & \multicolumn{1}{l|}{\textbf{GRCN}~\cite{yu2020graph}}      &   \url{https://github.com/PlusRoss/GRCN}  \\ \cline{3-4} 
\multicolumn{1}{l|}{}                                    & \multicolumn{1}{l|}{}                                     & \multicolumn{1}{l|}{RGCN~\cite{zhu2019robust}}      &  \url{https://github.com/ZW-ZHANG/RobustGCN}  \\ \cline{2-4} 
\multicolumn{1}{l|}{}                                    & \multicolumn{1}{l|}{\multirow{2}{*}{Direct Optimization}} & \multicolumn{1}{l|}{\textbf{ProGNN}~\cite{jin2020graph}}    & \url{https://github.com/ChandlerBang/Pro-GNN} \\ \cline{3-4} 
\multicolumn{1}{l|}{}                                    & \multicolumn{1}{l|}{}                                     & \multicolumn{1}{l|}{GSML~\cite{wan2021graph}}      & Code not available  \\ \hline
\multicolumn{1}{l|}{\multirow{2}{*}{Construction-based}} & \multicolumn{1}{l|}{Up-to-Bottom}                         & \multicolumn{1}{l|}{\textbf{PTDNet}~\cite{zheng2020robust}}    &   \url{https://github.com/flyingdoog/PTDNet}   \\ \cline{2-4} 
\multicolumn{1}{l|}{}                                    & \multicolumn{1}{l|}{Bottom-to-Up}                         & \multicolumn{1}{l|}{LDS~\cite{franceschi2019learning}}       &\url{https://github.com/lucfra/LDS-GNN} \\ \cline{1-4} 
\end{tabular}}
\end{sc}
		\end{small}
	\end{center}
\end{table*}

\begin{figure}[htb]
  \centering
  \includegraphics[width=0.45\textwidth]{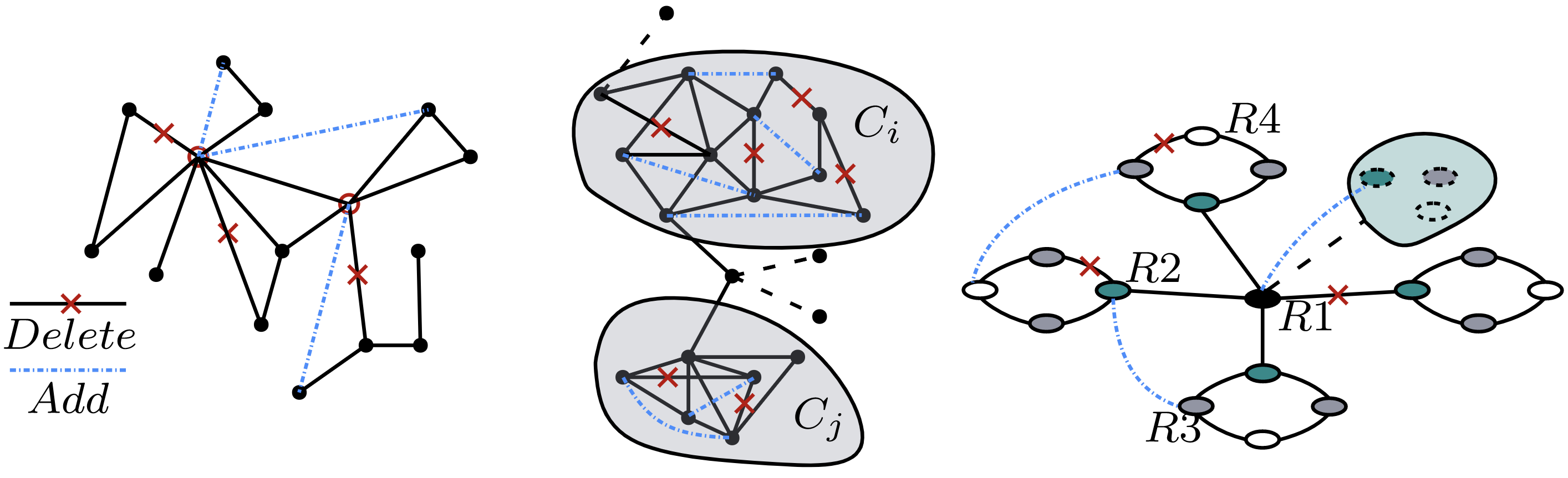}
  \caption{Demonstration for generating local(left), community(middle) and global(right) noise with flipping operation.}
  \label{fig:local_diagram}
  \vskip -0.1in
\end{figure}

\begin{figure*}[ht]
    \captionsetup[subfigure]{aboveskip=10pt}
    \centering
    \subfigure[Delete]{
    \label{Fig.sub.a}
    \includegraphics[width=0.32\textwidth]{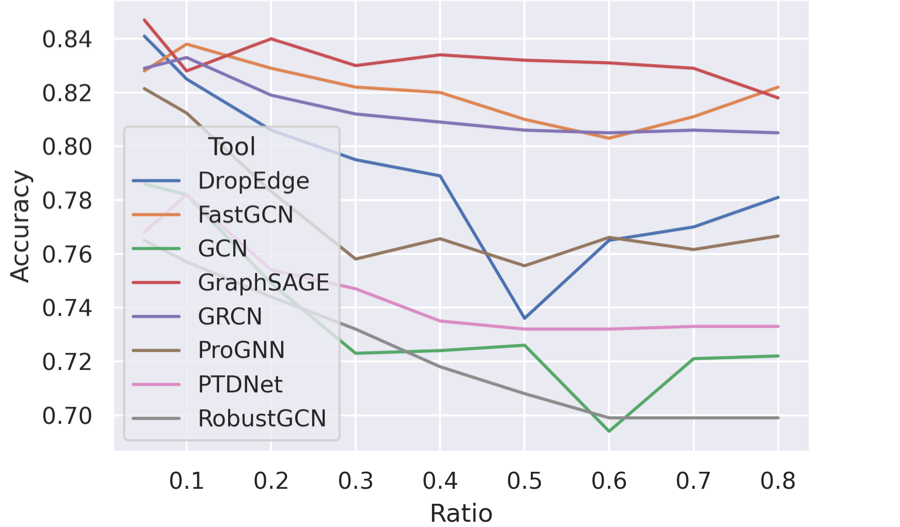}}
    \vspace{-0.25\baselineskip}
    \subfigure[Flip]{
    \label{Fig.sub.b}
    \includegraphics[width=0.32\textwidth]{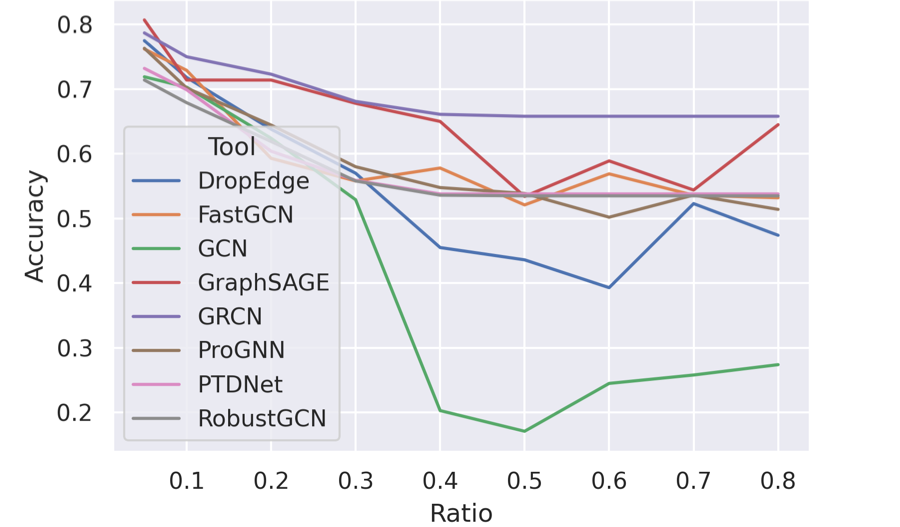}}
    \vspace{-0.25\baselineskip}
    \subfigure[Add]{
    \label{Fig.sub.c}
    \includegraphics[width=0.32\textwidth]{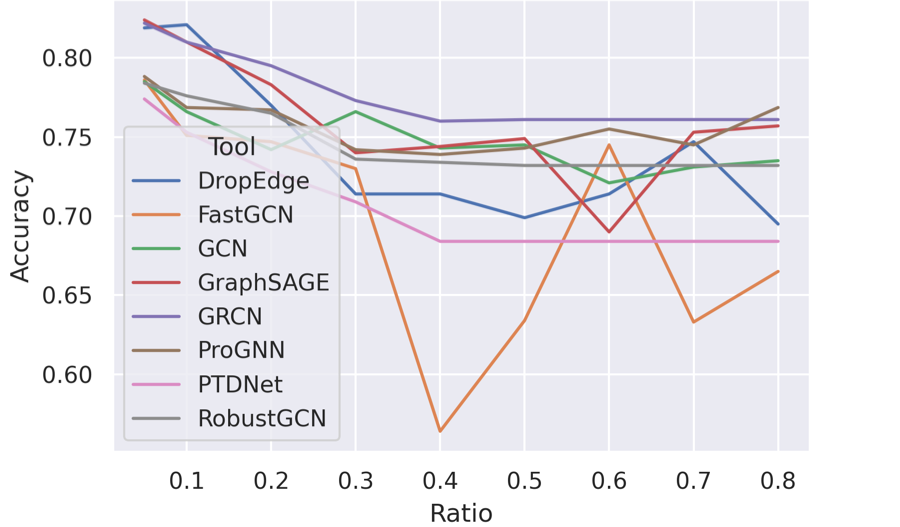}}
    \vspace{-0.25\baselineskip}
    \\
    \subfigure[Delete]{
    \label{Fig.sub.d}
    \includegraphics[width=0.32\textwidth]{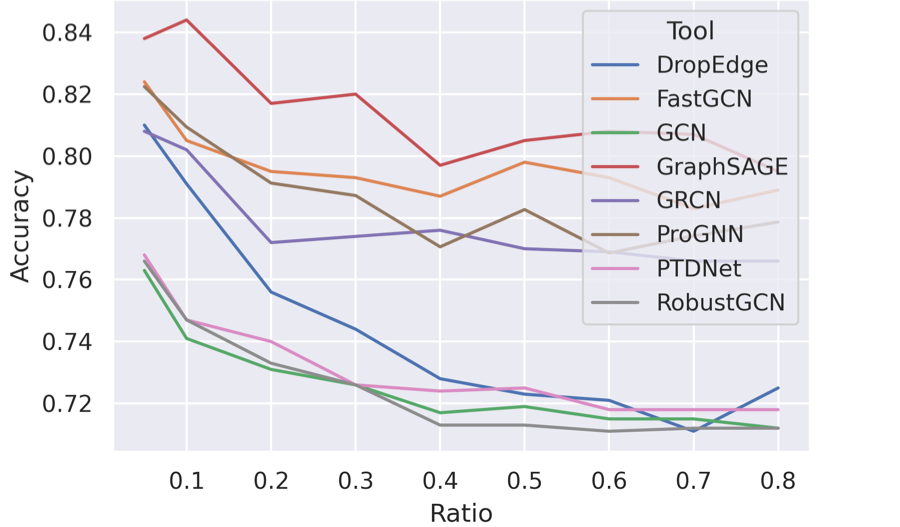}}
    \vspace{-0.25\baselineskip}
    \subfigure[Flip]{
    \label{Fig.sub.e}
    \includegraphics[width=0.32\textwidth]{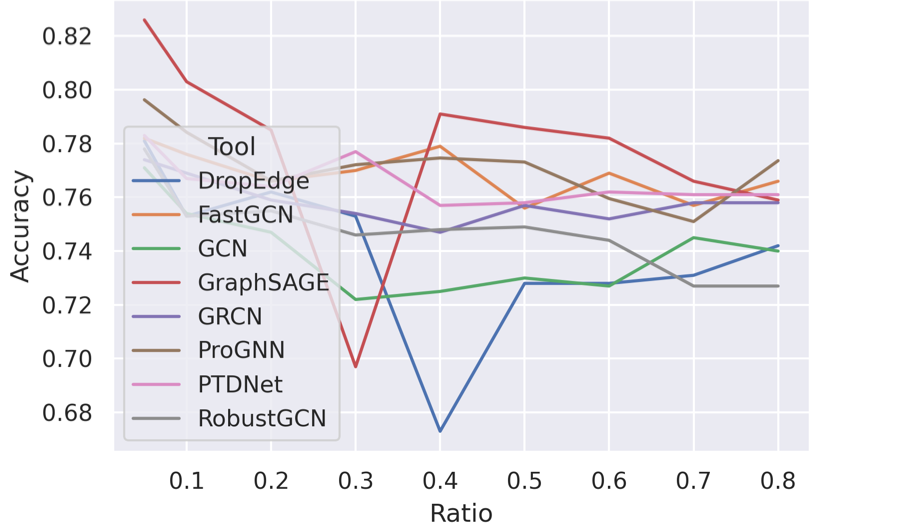}}
    \vspace{-0.25\baselineskip}
    \subfigure[Add]{
    \label{Fig.sub.f}
    \includegraphics[width=0.32\textwidth]{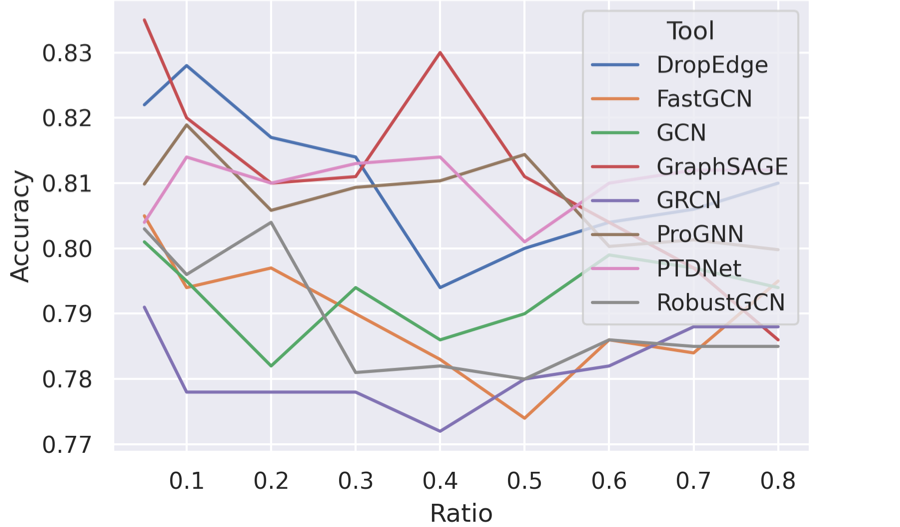}}
    \vspace{-0.25\baselineskip}
    \\
    \subfigure[Delete]{
    \label{Fig.sub.g}
    \includegraphics[width=0.32\textwidth]{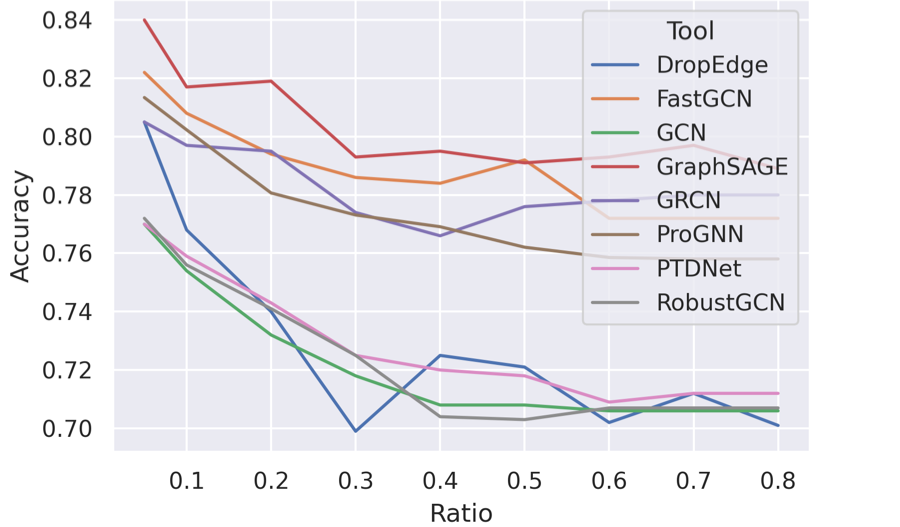}}
    \subfigure[Flip]{
    \label{Fig.sub.h}
    \includegraphics[width=0.32\textwidth]{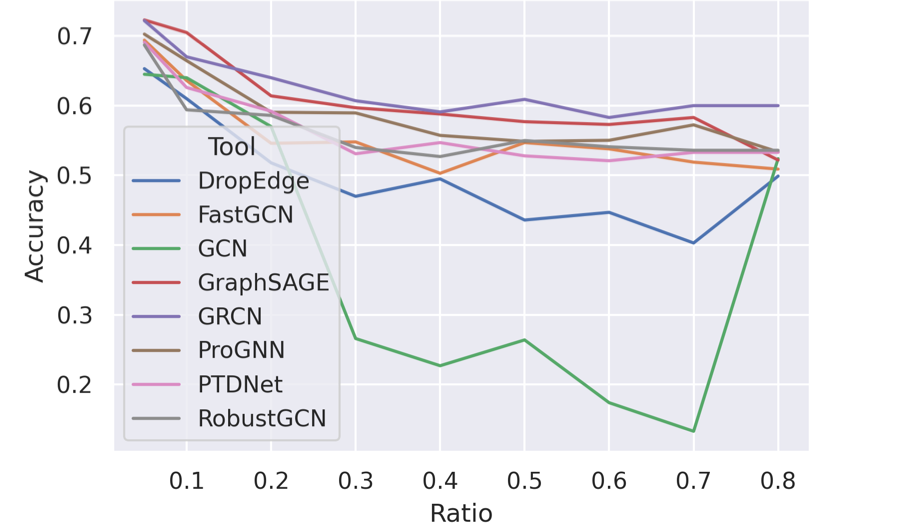}}
    \subfigure[Add]{
    \label{Fig.sub.i}
    \includegraphics[width=0.32\textwidth]{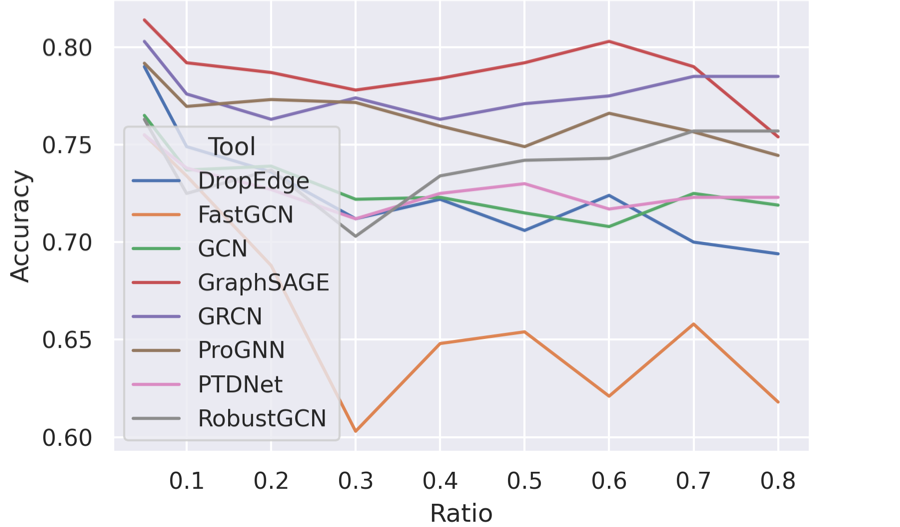}}\\
    \caption{\label{overall}  Overall performance of different GNNs to different structural noises. (a)-(c) are local noises, and (d)-(f) are community noises, and (g)-(i) are global noises.}
    \vskip -0.2in
\end{figure*}

\section{Noise Generation}
To compare the performance of existing robust GNNs under different granularity levels of structural noises, we propose three methods to inject noises in the graph structure. These three methods are designed from a local, community, and global perspective over the graph structure. Thus, we distinguish these generated noises as \textit{local noise}, \textit{community noise}, and \textit{global noise}, respectively. An illustration of generating these three kinds of noise is shown in Figure~\ref{fig:local_diagram}. Moreover, there are three basic operations for each type of noise: deleting, flipping, and adding edges. Due to the page limit, we will fully introduce these three basic operations only in the subsection of local noise.

\subsection{Local Noise}
The purpose of introducing local noise is to control the perturbation on graph structure at a node level. With this initiative, noise is generated in the following manner: firstly, we choose a suitable metric for measuring the importance of each node in the original graph, e.g., the \textit{degree} or the \textit{betweenness centrality}. We provide an implementation using the degree for simplicity. Afterward, all nodes are sorted based on the magnitude of their degrees descendingly. Finally, we can manipulate the graph structure using three basic operations introduced above based on the sorted nodes and their corresponding neighbors. In specific, we have
\begin{itemize}
    \item \textit{Deleting Edges}. The strategy of deleting edges is to set a threshold to limit the information flow over a node in the graph. For those nodes whose degrees exceed the threshold value, we randomly delete the edges connected to them until their new degrees are equal to the threshold. (We observe the inference accuracy of most GNN models declines severely when the threshold value is the mode of degrees.) After deleting edges on the original graph, a certain number of nodes that initially have the largest degrees will be degraded to the threshold level. 
    \item \textit{Flipping Edges}. Flipping edges is a more complicated operation, while it means replacing the existing edges with artifact ones. The commonplace between flipping and deleting edges is they both modify nodes that have the largest degrees. For each modified node, we repeat this process a certain number of times (specified later): Randomly connecting it with another node except its original neighbors, after randomly deleting an existing edge. Note that a golden standard we should comply with is the number of edges after flipping edges remains the same as the original graph, and all added edges cannot appear in the original graph.  This standard also applies to community noise and global noise. 
    \item \textit{Adding Edges}. The manipulating objects of adding edges are consistent with the previous two operations. After determining the modified node, we randomly add a certain number of edges for that node.  
\end{itemize}

To facilitate the comparison among operations, we specify the amount of flipped or added edges to be the same as that of deleted edges for each manipulated node. Besides, we introduce a ratio to determine the number of nodes to be modified for all operations. 

Technically, the process of adding noise is adding a mask to the original graph structure. While the local noise remains at a node level, deleting/flipping/adding edges can be considered as filtering/shuffling/expanding the information flow over a couple of nodes during the message passing process \cite{lanchantin2019neural}. 

\subsection{Community Noise}
Rather than merely influencing a certain number of nodes in the graph, we expect to expand the perturbing scope on the graph that noise makes an larger impact. To achieve this goal, we refer to the community~\cite{fortunato2010community} to introduce noises into the original graph. Accordingly, we sketch the following process for noise generation: initially, the input graph is grouped into $k$ communities ($k$ is a hyperparameter), while each community is densely connected internally. We apply the Louvain algorithm~\cite{blondel2008fast} to detect communities for simplicity. Then, we can conduct different operations based on the grouping result. 
\begin{itemize}
    \item \textit{Flipping Edges}. Since the grouping result is each node in the graph will be assigned to one community, we can use this information to categorize all edges into two classes: within a community and between communities. For edges within a community, we sample a certain ratio of edges and delete them from the edge set. Next, we randomly add a set of edges whose both endpoints are within this community. For edges between communities, we also delete existing edges at first. However, when adding a new edge, we should make sure endpoints of the edge originate from two different communities. 
\end{itemize}
The operation object when generating noise is under the community basis, thus, this kind of perturbation spreads the influence of noise on a community/subgraph level. 

\subsection{Global Noise}
Furthermore, we extend the perturbing scope of noise to a global level. To reach this goal, a way to capture the structural properties of nodes from a global perspective is desired. In social science and graph mining, \textit{role}~\cite{rossi2014role} is a straightforward concept to mine graph structures globally. Roles can represent node-level connectivity patterns, e.g., bridge, cliquey, isolated, which are shared globally on a graph. In practice, we employ RolX \cite{henderson2012rolx} to cluster nodes with similar structural features into the same group. After obtaining the role assignment of each node, we can undertake operations on the graph. 
\begin{itemize}
    \item \textit{Flipping Edges}. RolX clusters nodes based on globally structural similarity, so nodes that have the same role may be scattered in various positions of the graph. Thus, most edges have endpoints with different roles. We can use role information to categorize all edges into two classes: within the role set and between role sets. The processing for edges within the role set follows the same pattern as Community Noise. Differently, for edges between role sets, whenever we randomly delete an edge, we add a new one that has the same endpoints as the original one. 
\end{itemize}
While the noise is mainly made universally in the graph, we call the perturbation is propagated to a global level.

\section{Experiments}
We conduct extensive experiments to evaluate difference robust GNNs under three granularity levels of noises we proposed. The implementation of generating the structural noise is available\footnote{\url{https://github.com/Jantory/GNN-Comparison}}. Additionally, we extend this to a framework that can be easily adapted to make comparison for different robust GNNs under any noise settings. 

The experiments apply two benchmark datasets that are commonly used for evaluating GNNs, i.e., Cora and Citeseer~\cite{sen2008collective}. The data split for training/validation/test set originates from \cite{yang2016revisiting}. We sequentially generate local noise, community noise, and global noise to the graph structure. Firstly, the structural noise is generated by a specific ratio ranging from 0.05 to 0.8 with the interval 0.05 for each type of noise. Afterward, we fed the noised graph to eight models for training and testing of the semi-supervised classification task. Because the noise generation and training for a few methods depend on the random process, we conduct the same experiment six times (by fixing random seed) to avoid variance caused by accidents or other factors.

\begin{figure}[htb]
  \centering
  \includegraphics[width=0.49\textwidth]{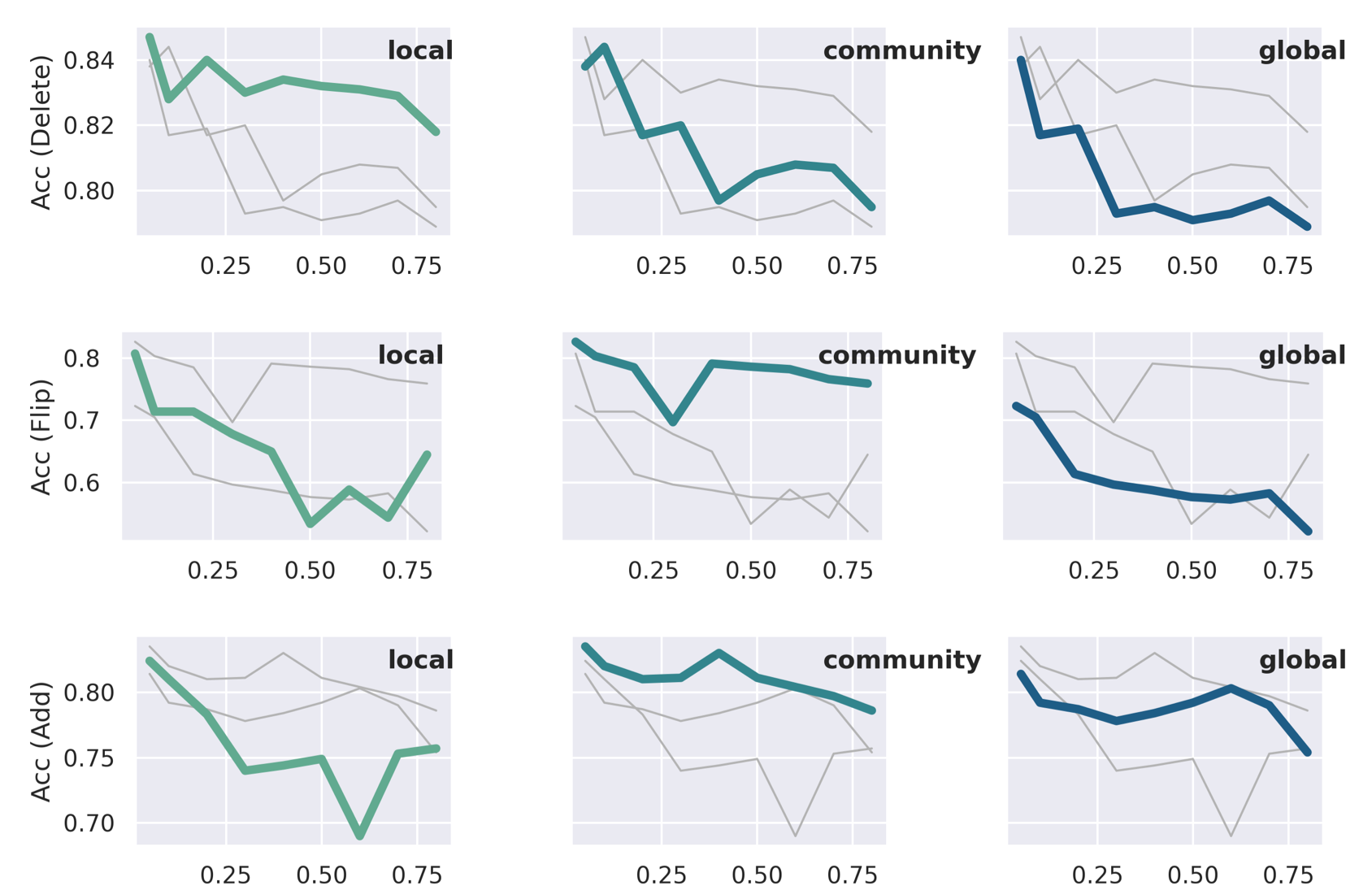}
  \caption{GraphSAGE.}
  \label{fig:graphsage}
  \vskip -0.1in
\end{figure}
\begin{figure}[htb]
  \centering
  \includegraphics[width=0.49\textwidth]{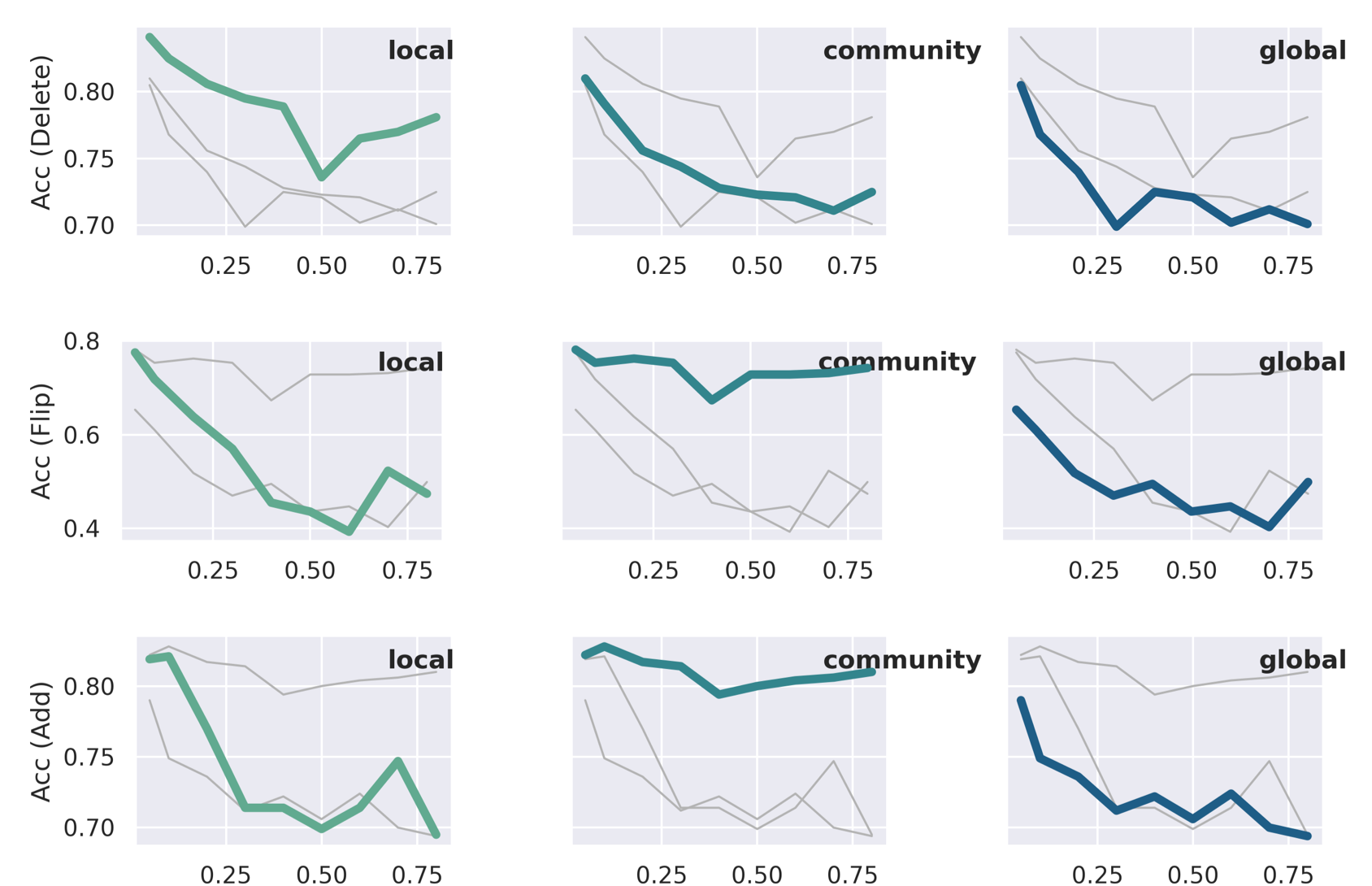}
  \caption{DropEdge.}
  \label{fig:dropedge}
  \vskip -0.2in
\end{figure}
\begin{figure}[htb]
  \centering
  \includegraphics[width=0.49\textwidth]{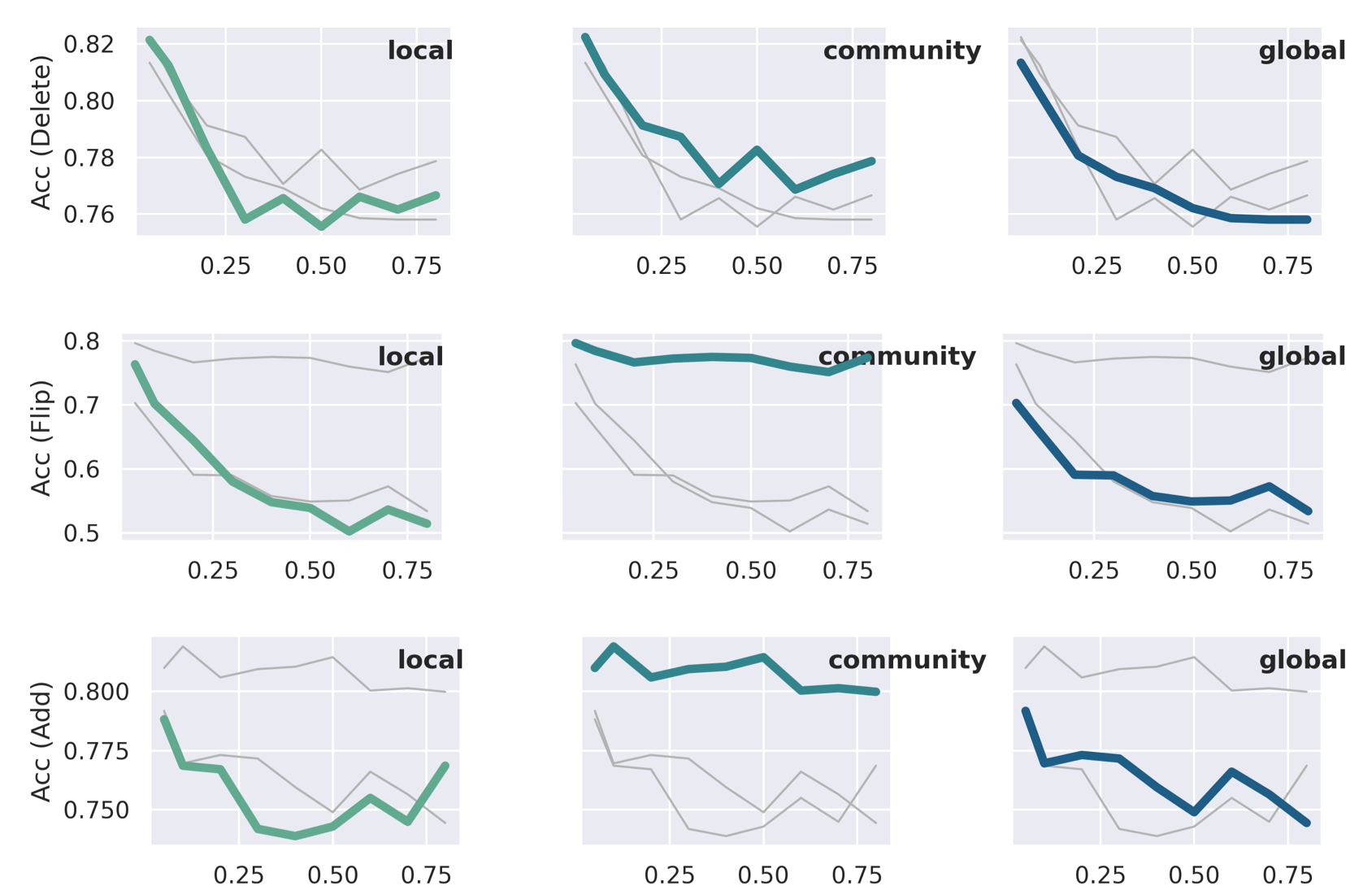}
  \caption{Pro-GNN.}
  \label{fig:prognn}
  \vskip -0.2in
\end{figure}

\textbf{Results}. The overall performance, i.e., classification accuracy, of different robust GNNs under different levels of structural noises is shown in Figure~\ref{overall}. To be more specific, we highlight some models in Figures~\ref{fig:graphsage} to~\ref{fig:gcn} and provide the full results in the supplementary materials. From the results, some conclusions can be drawn:
\begin{itemize}
    \item Overall, while node-sampling methods outperform other models in several cases, revision-based methods are stable under most noise settings. For instance, GraphSAGE achieves the best performance under most noise settings. However, as shown in Figure~\ref{overall}(e) and (f), it suffers a sharp fluctuation in exceptional cases. The possible reason that leads to this phenomenon is sampling process, and this situation happens in all sample-based methods. Besides, GraphSAGE achieves inferior performance when the graph contains isolated nodes (This can be observed from the empirical results on Citeseer, detailed in the complementary materials). Moreover, although the revision-based methods only have moderate performance, they are stable enough to maintain a gentle and smooth performance curve under any noises, e.g., Pro-GNN. In particular, when the graph is severely noised, GRCN obtains the best performance shown in Figure~\ref{overall}(b), (c).
    \item 
    % under different levels of noises, which ones are better
    In Figures~\ref{fig:graphsage}-\ref{fig:prognn}, it can be observed that global noise will degenerate the model performance most for almost all tools. Nevertheless, different types of noise show two distinct characteristics under three basic operations. When we delete edges from the graph, local noise has the least impact on model performance, followed by community noise and then global noise. However, if the operations conducted on the graph are flipping edges and adding edges, community noise has the most negligible impact on model performance, followed by local and then global noise. Note that community noise only yields a trivial effect on model performance. It demonstrates that models are easier to capture community information and show more robustness towards community noises.
    \item 
    % under different strategy, which ones are better
    Flipping edges greatly affects performance of the model compared to others. We can also find the following trends: Firstly, node-sampling methods like GraphSAGE and FastGCN are mostly suitable for deleting operations (see from Figure \ref{overall}(a)(d)(g)); Revision-based methods such as GRCN and Pro-GNN, and construction-based methods like PTDNet have the best and stablest performance for flipping operation (see from Figure \ref{overall}(b)(e)(h)); Edge-sampling methods and revision-based methods are more capable of handling adding operation especially when the graph is not severely noised (see from Figure \ref{overall}(c)(f)(i)).
    
\end{itemize}

\section{Discussions}
% \textbf{Discussions contain practical suggestions when selecting suitable GNNs. Some tips include: if no information about noise, which should select? If we have clue about structural noises, which should select?}

We have categorized structural noise into different levels, and three levels of noise are designed correspondingly to evaluate the robustness of GNNs. The subsequent experiments reveal the performance of different robust GNNs under several noise settings. These comparative results helps select models in the future. Overall, when no information about noise is provided, node-sampling models can be a reasonable choice for higher performance, while revision-based models should always be used as an alternative to yield stable and robust performance. 

On the other hand, if we have a clue about structural noises, we need to choose models based on noise characteristics. Firstly, the more global noise is, the less we should rely on sample-based models. Secondly, node-sampling methods are more suitable for deleting operations; revision-based and construction-based methods are ideal for flipping operation; and revision-based and edge-sampling methods are more suitable for flipping operation.

\section{Conclusion}
In this work, we conducted a comprehensive and systematical comparative study on different types of robust GNNs under consistent structural noise settings. To make a fair comparison, two aspects including models and noises have been considered: from the noise aspect, we design three different levels of structural noises, i.e., local, community, and global noises. From the model aspect, we select some representative models from sample-based, revision-based, and construction-based robust GNNs. Based on the experimental results, some practical suggestions have been provided when selecting suitable GNNs towards structural noises. Some future directions include designing some measure to detect noise type and proposing robust GNNs for general noises.

% \newpage
% \begin{figure*}
%     \centering
%     \subfigure[GraphSAGE]{
%     \label{Fig.sub.1}
%     \includegraphics[width=0.48\textwidth]{figures/GraphSAGE_Mode.png}}
%     \subfigure[DropEdge]{
%     \label{Fig.sub.2}
%     \includegraphics[width=0.48\textwidth]{figures/Dropedge_Mode.png}}\\
%     \subfigure[Pro-GNN]{
%     \label{Fig.sub.1}
%     \includegraphics[width=0.48\textwidth]{figures/ProGNN_Mode.png}}
%     \subfigure[GCN]{
%     \label{Fig.sub.2}
%     \includegraphics[width=0.48\textwidth]{figures/GCN_Mode.png}}\\
%     \caption{\label{Fig.Role transition}  Sankey diagram of nodes and roles evolution in five timestep with different methods. }
% \end{figure*}

% Use \bibliography{yourbibfile} instead or the References section will not appear in your paper
%\input{aaai22.bbl}
\bibliography{aaai22}

\newpage
\section{Complementary Materials}

\subsection{Experimental Settings}
We conduct extensive experiments to evaluate difference robust GNNs under three noise levels we proposed. We assume that the hyperparameters and parameters of all models have been tuned to achieve the best performance in the provided code, so our comparison did not tune them. For community noise, we utilize python-louvian\footnote{\url{https://github.com/taynaud/python-louvain}} module to find the best community partition. For global noise, GraphRole\footnote{\url{https://github.com/dkaslovsky/GraphRole}} is applied while six is default set as the number of roles. The implementation of generating three different level of structural noise is available\footnote{\url{https://github.com/Jantory/GNN-Comparison}}. Additionally, we extend this to a framework that can be easily adapted to make comparison for different robust GNN models under any noise settings. 

\subsection{Datasets}
The experiments apply two standard benchmark datasets that are commonly used for evaluating graph-based learning tasks, i.e., Cora and Citeseer. Precisely, experiments on the Cora dataset are used to make several assumptions; then, we conduct the same work on Citeseer for revising and validating them. The data split for training/validation/test set originates from \cite{yang2016revisiting}. Basic statistics are shown in Table \ref{dataset}. 

\subsection{Computing Resources}
The GPU resource we use is 8x GeForce GTX 1080 Ti (RAM, 3584 CUDA cores), and the CPU is 2x Intel Xeon Broadwell-EP 2683v4 @ 2.1GHz (64 hyperthreads). We exclusively perform the training and evaluation of GNNs on GPU, but generating noise on graph structure is on CPU.

\subsection{Experimental Results}
\begin{figure}[htb]
  \centering
  \includegraphics[width=0.49\textwidth]{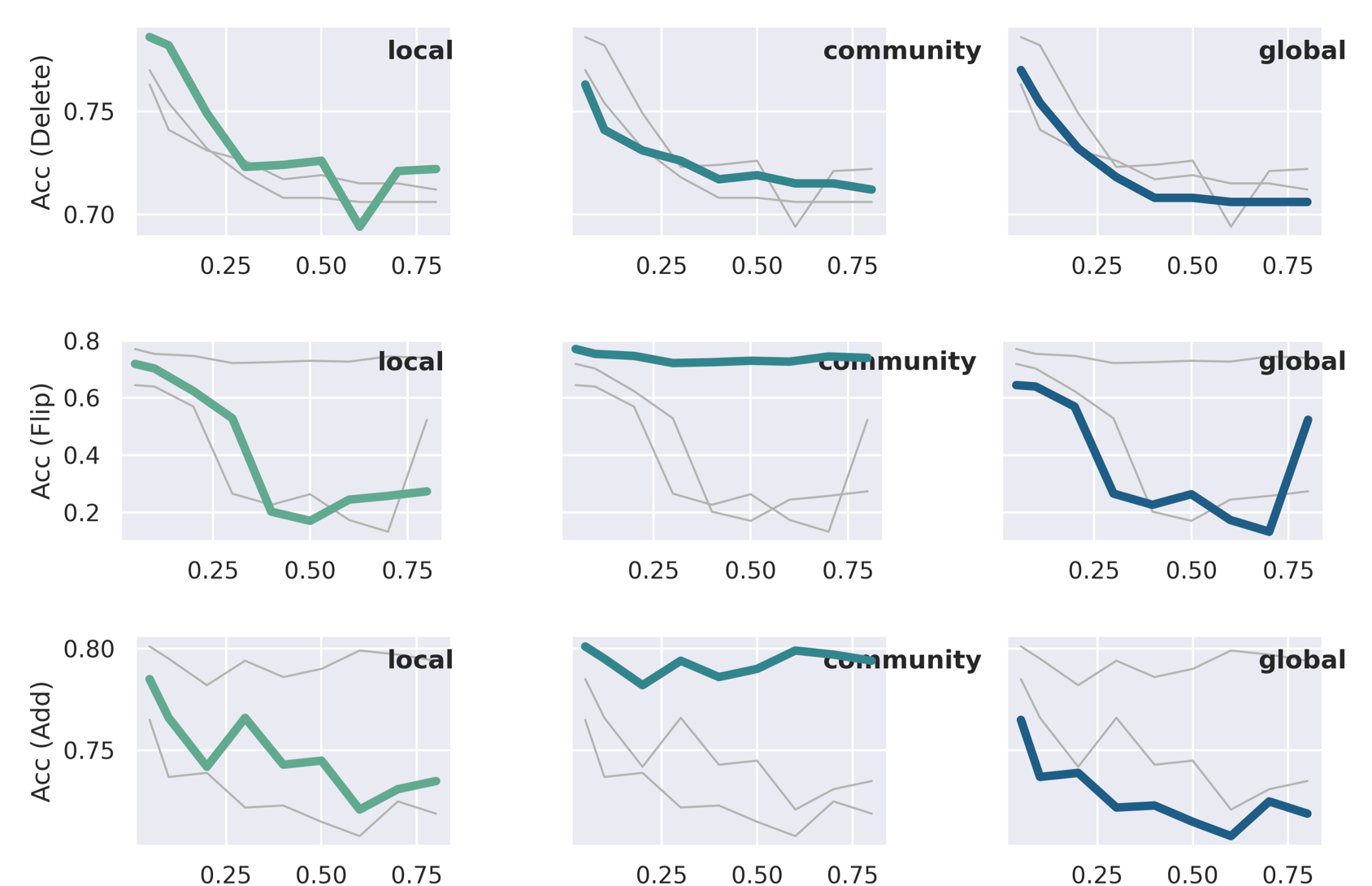}
  \caption{GCN.}
  \label{fig:gcn}
  \vskip -0.2in
\end{figure}

\begin{figure}[htb]
  \centering
  \includegraphics[width=0.49\textwidth]{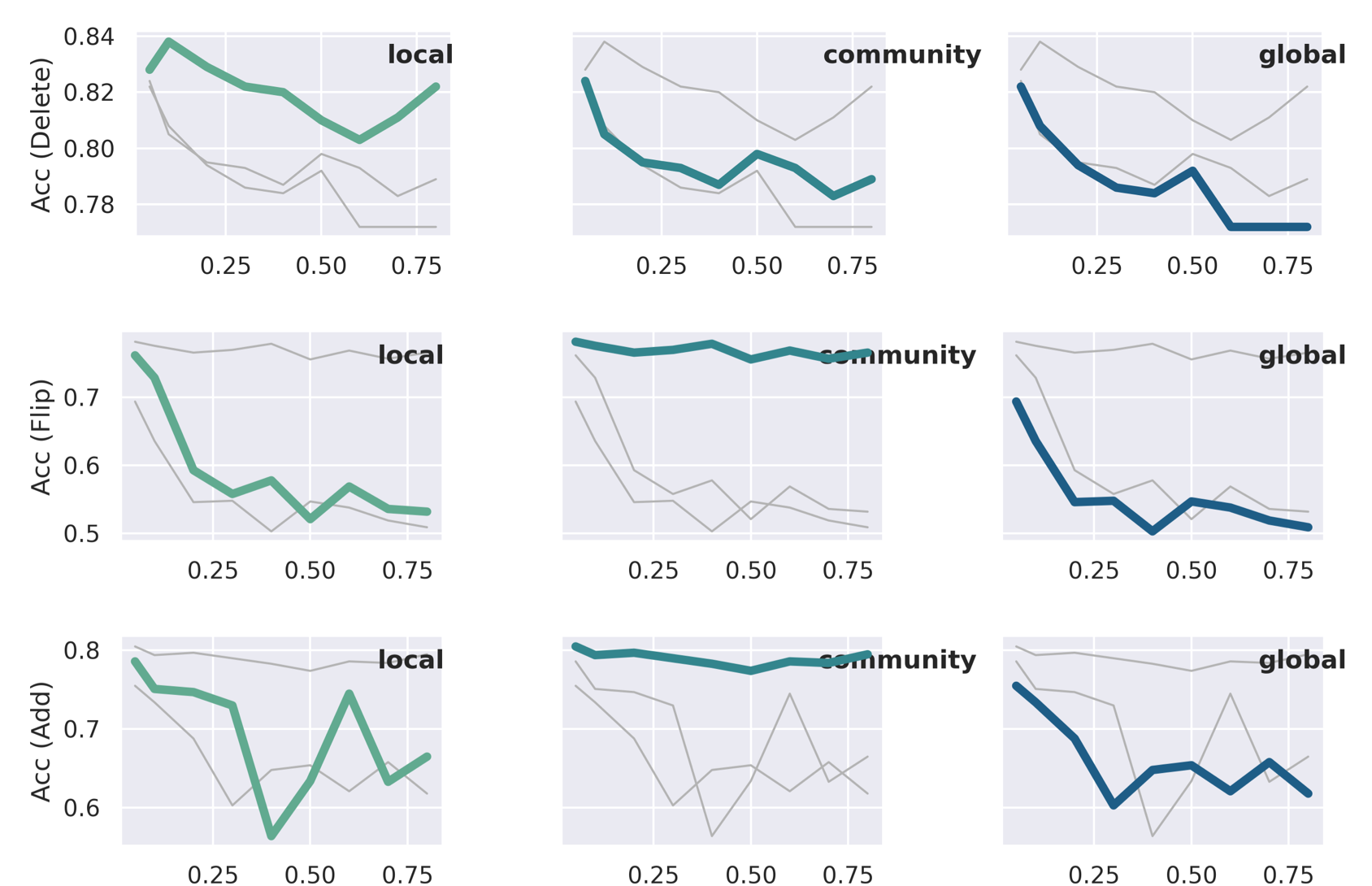}
  \caption{FastGCN.}
  \label{fig:fastgcn}
  \vskip -0.2in
\end{figure}

\begin{figure}[htb]
  \centering
  \includegraphics[width=0.49\textwidth]{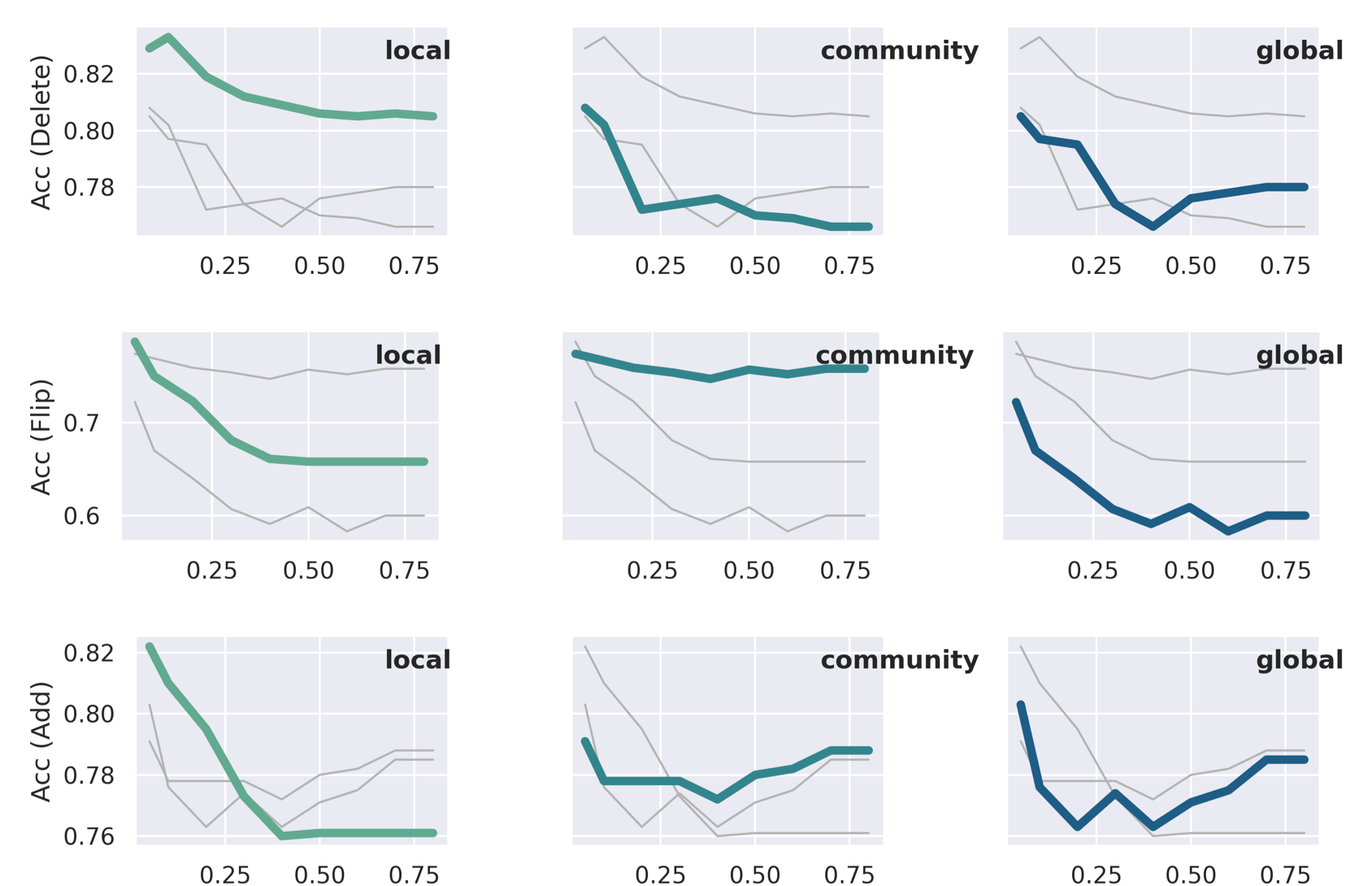}
  \caption{GRCN.}
  \label{fig:grcn}
  \vskip -0.2in
\end{figure}

\begin{figure}[htbp]
  \centering
  \includegraphics[width=0.49\textwidth]{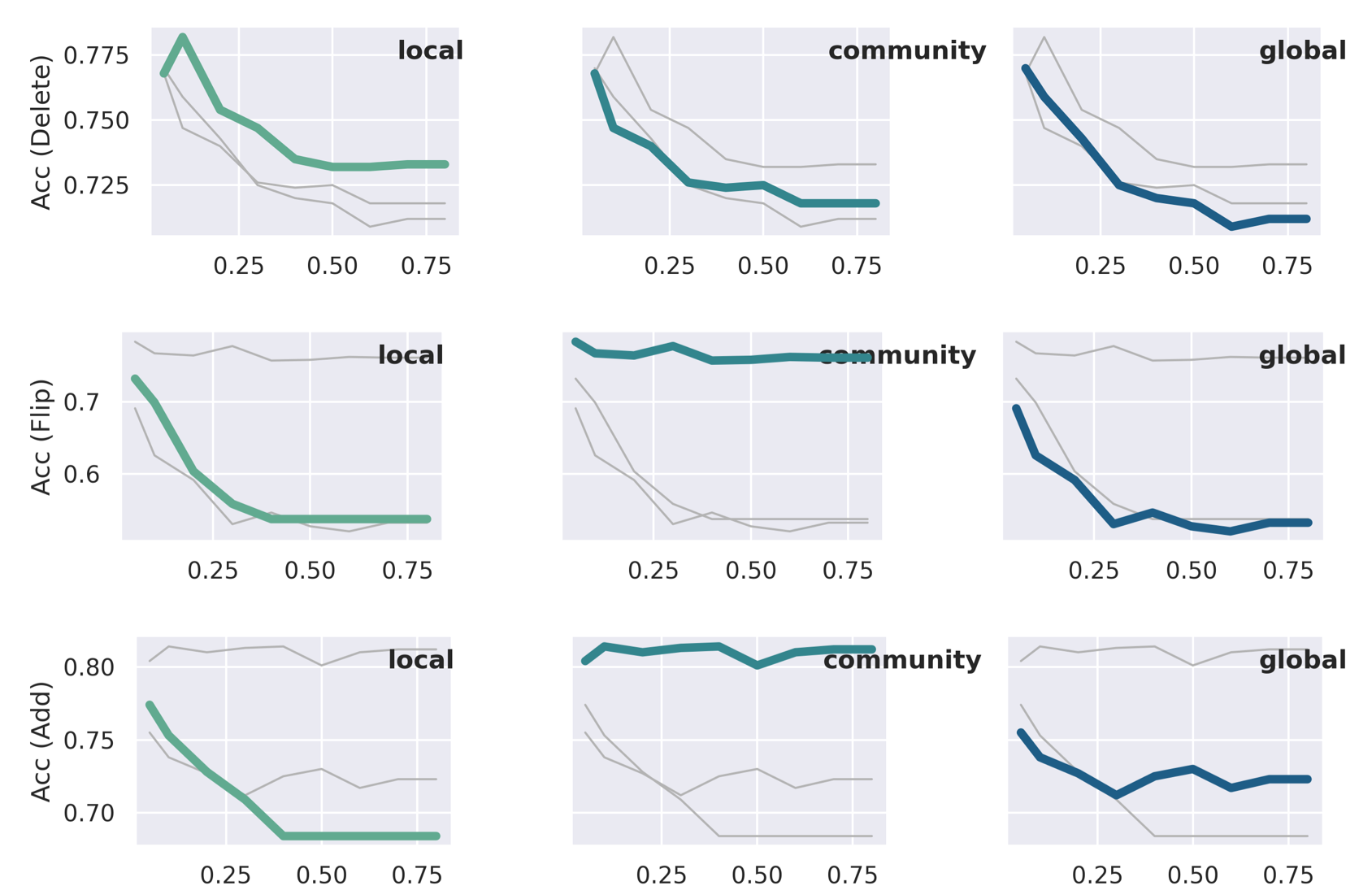}
  \caption{PTDNet.}
  \label{fig:ptdnet}
  \vskip -0.2in
\end{figure}

\begin{figure}[htb]
  \centering
  \includegraphics[width=0.49\textwidth]{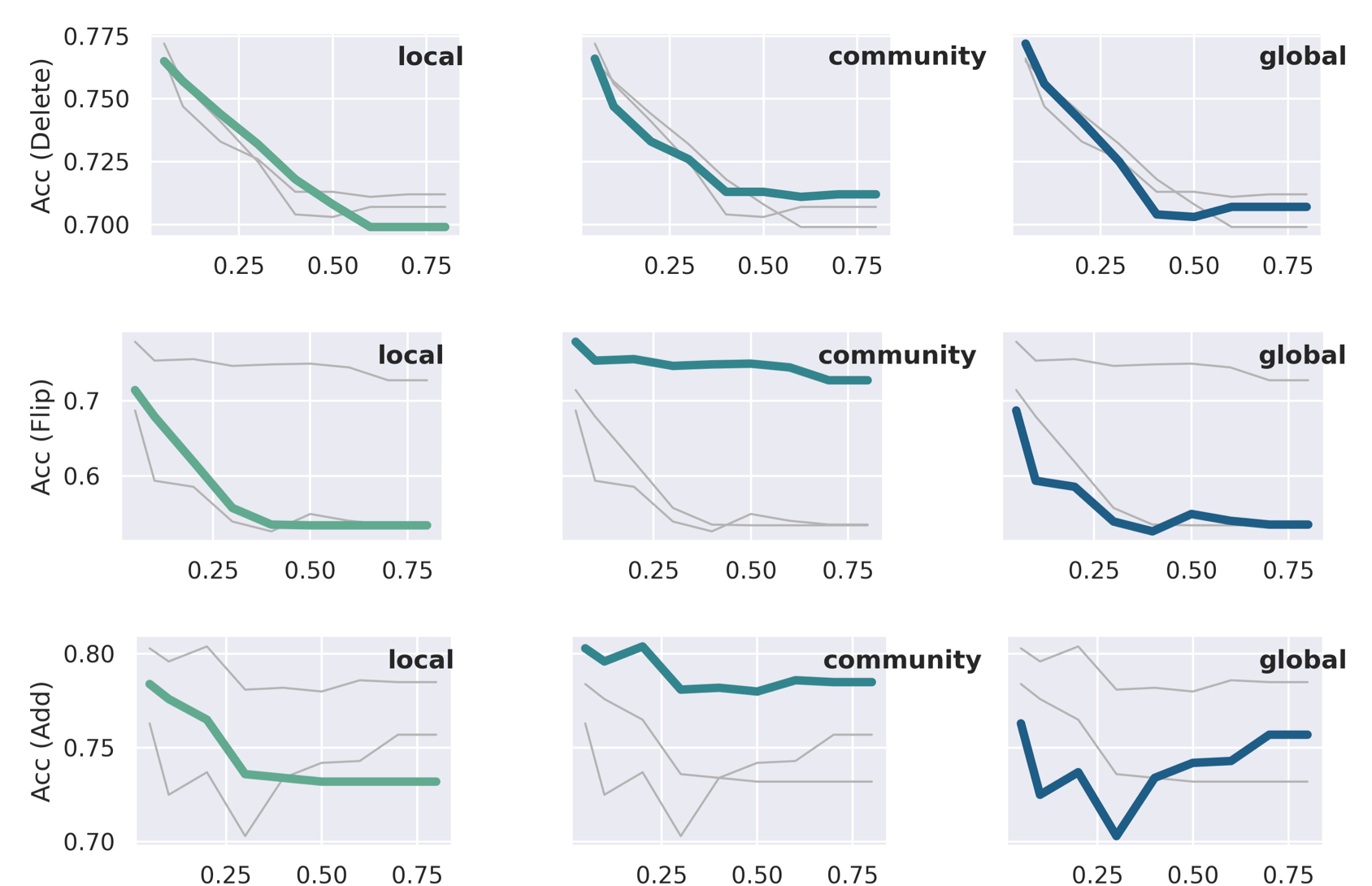}
  \caption{RobustGCN.}
  \label{fig:robustgcn}
  \vskip -0.2in
\end{figure}

\begin{figure*}[ht]
    \captionsetup[subfigure]{aboveskip=10pt}
    \centering
    \subfigure[Delete]{
    \label{Fig.sub.1.1}
    \includegraphics[width=0.32\textwidth]{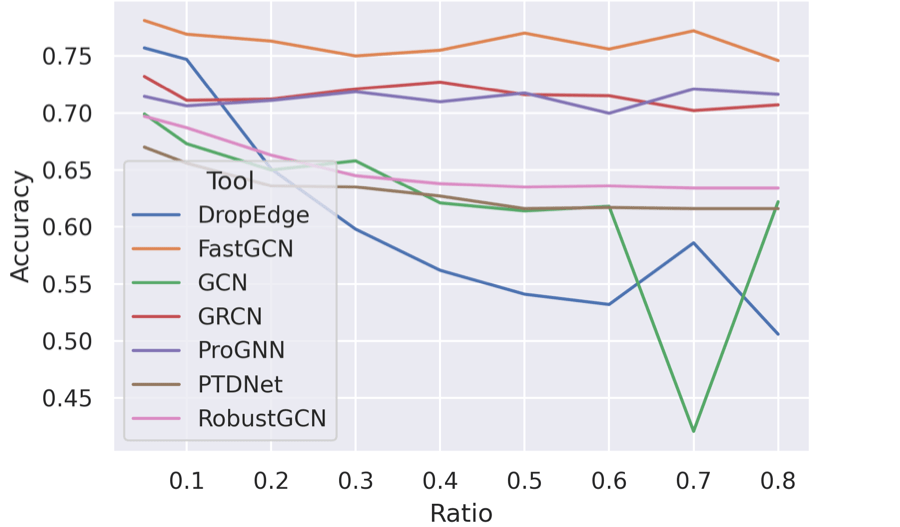}}
    \vspace{-0.25\baselineskip}
    \subfigure[Flip]{
    \label{Fig.sub.2.1}
    \includegraphics[width=0.32\textwidth]{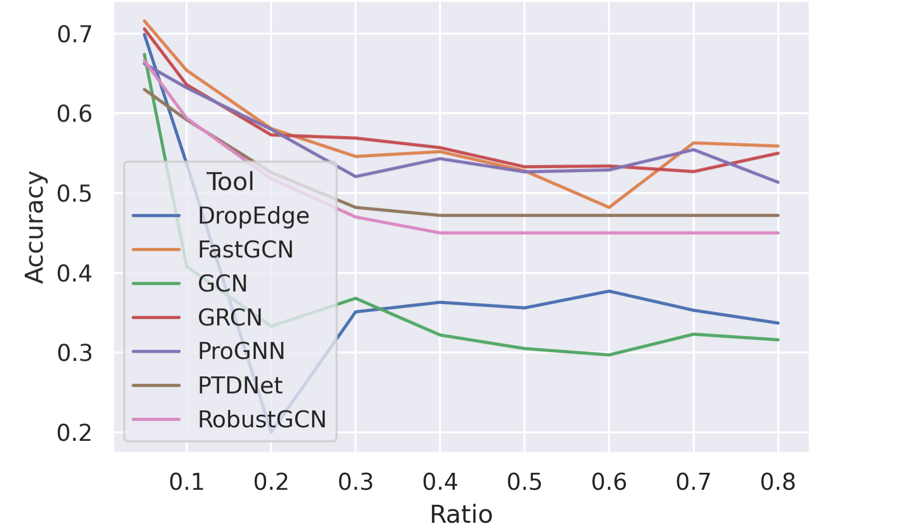}}
    \vspace{-0.25\baselineskip}
    \subfigure[Add]{
    \label{Fig.sub.3.1}
    \includegraphics[width=0.32\textwidth]{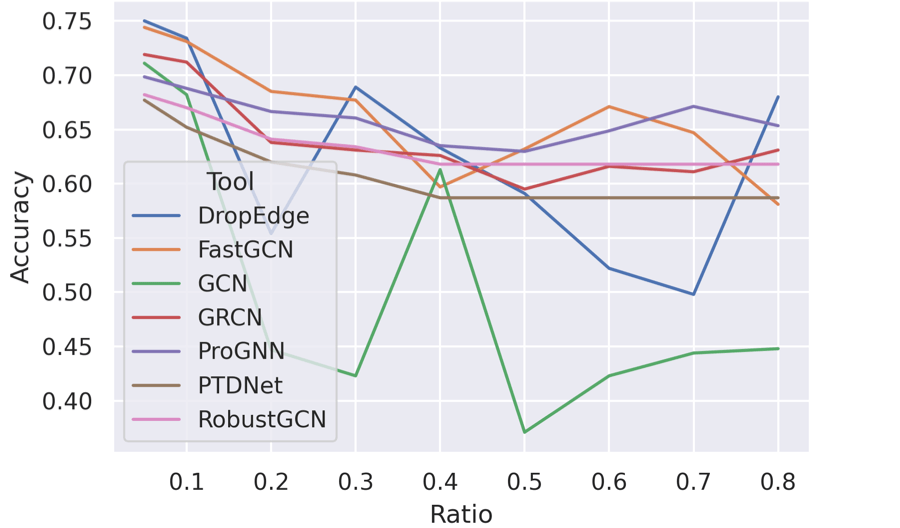}}
    \vspace{-0.25\baselineskip}
    \\
    \subfigure[Delete]{
    \label{Fig.sub.1.2}
    \includegraphics[width=0.32\textwidth]{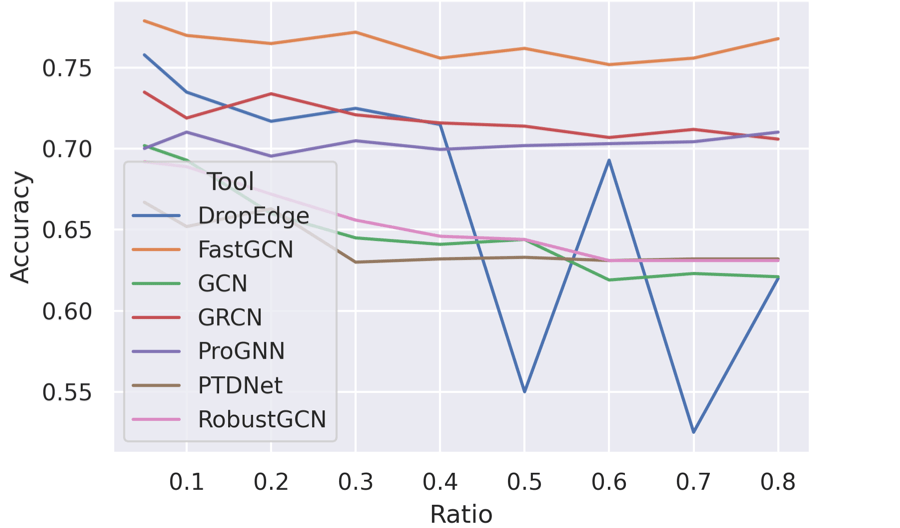}}
    \vspace{-0.25\baselineskip}
    \subfigure[Flip]{
    \label{Fig.sub.2.2}
    \includegraphics[width=0.32\textwidth]{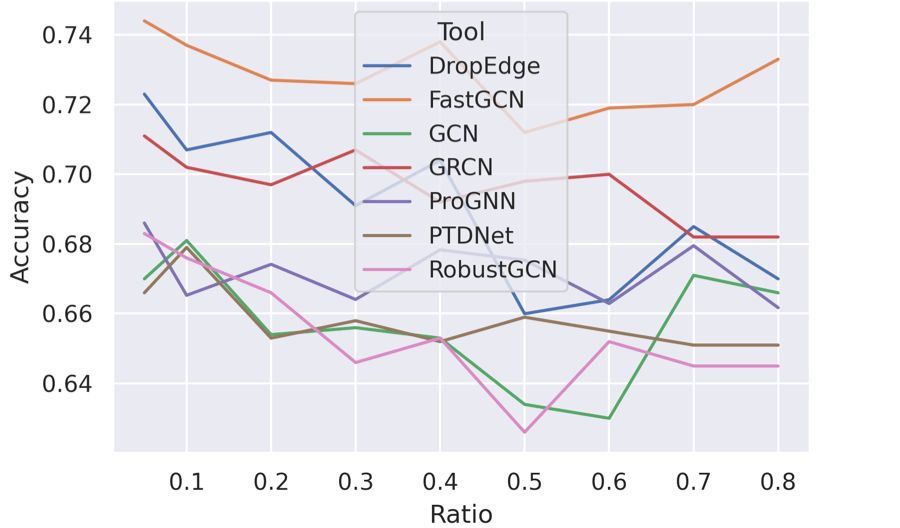}}
    \vspace{-0.25\baselineskip}
    \subfigure[Add]{
    \label{Fig.sub.3.2}
    \includegraphics[width=0.32\textwidth]{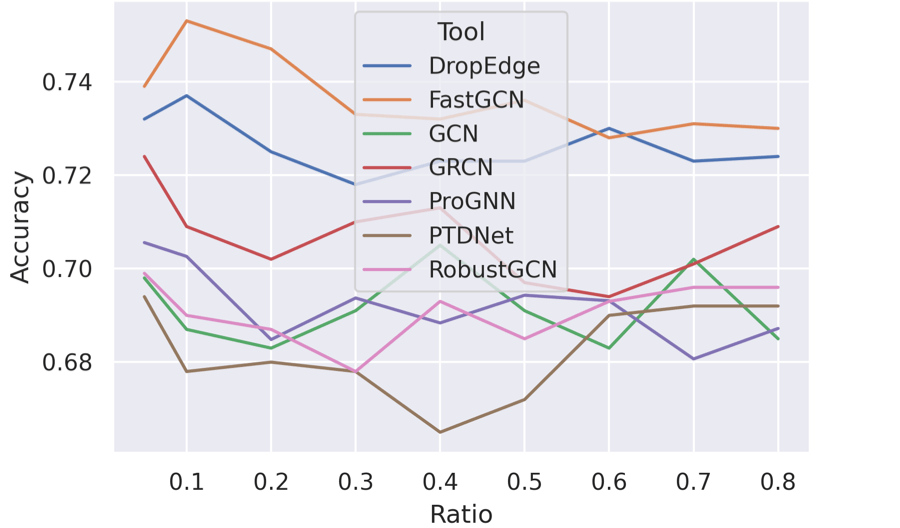}}
    \vspace{-0.25\baselineskip}
    \\
    \subfigure[Delete]{
    \label{Fig.sub.1.3}
    \includegraphics[width=0.32\textwidth]{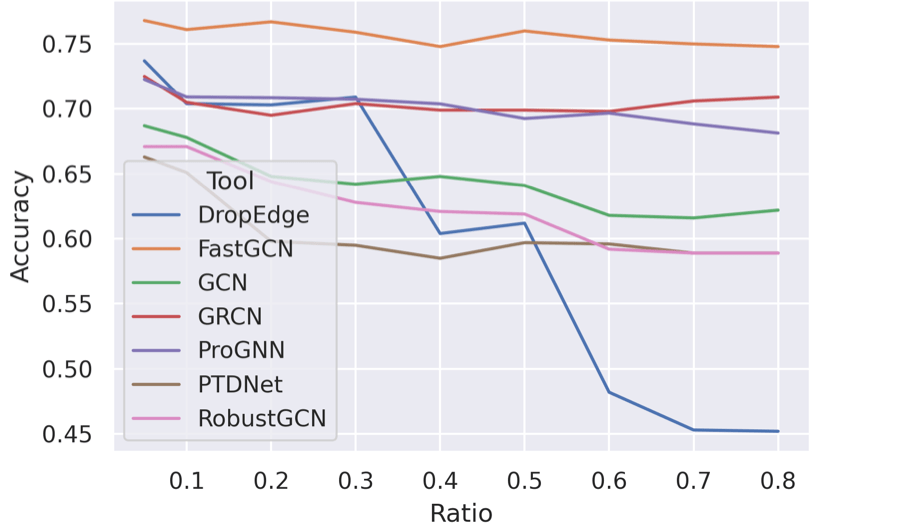}}
    \subfigure[Flip]{
    \label{Fig.sub.2.3}
    \includegraphics[width=0.32\textwidth]{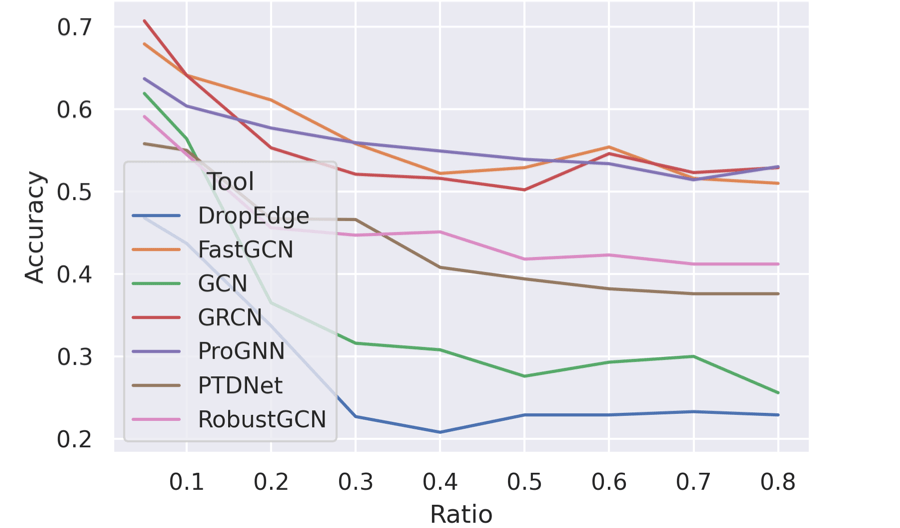}}
    \subfigure[Add]{
    \label{Fig.sub.3.3}
    \includegraphics[width=0.32\textwidth]{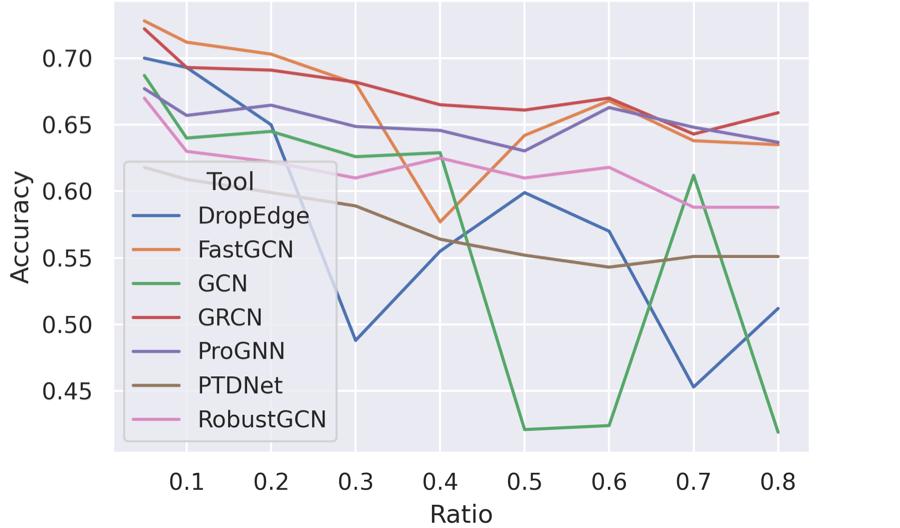}}\\
    \caption{\label{overall-citeseer}  Overall performance of different GNNs (without GraphSAGE) to different structural noises on Citeseer. (a)-(c) are local noises, and (d)-(f) are community noises, and (g)-(i) are global noises.}
\end{figure*}

\begin{figure}[htb]
  \centering
  \includegraphics[width=0.49\textwidth]{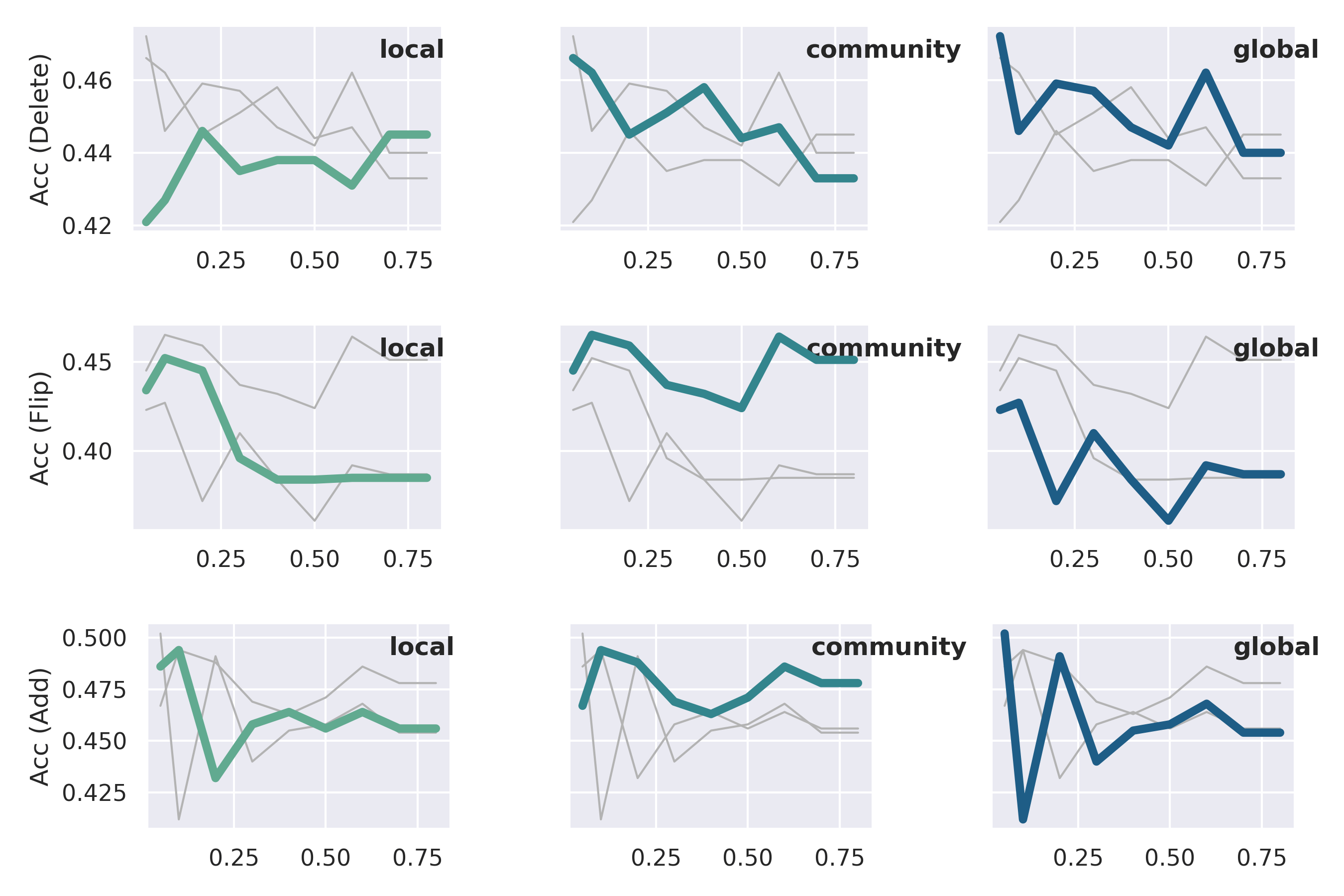}
  \caption{GraphSAGE on Citeseer.}
  \label{fig:graphsage_c}
  \vskip -0.2in
\end{figure}

\begin{figure}[htb]
  \centering
  \includegraphics[width=0.49\textwidth]{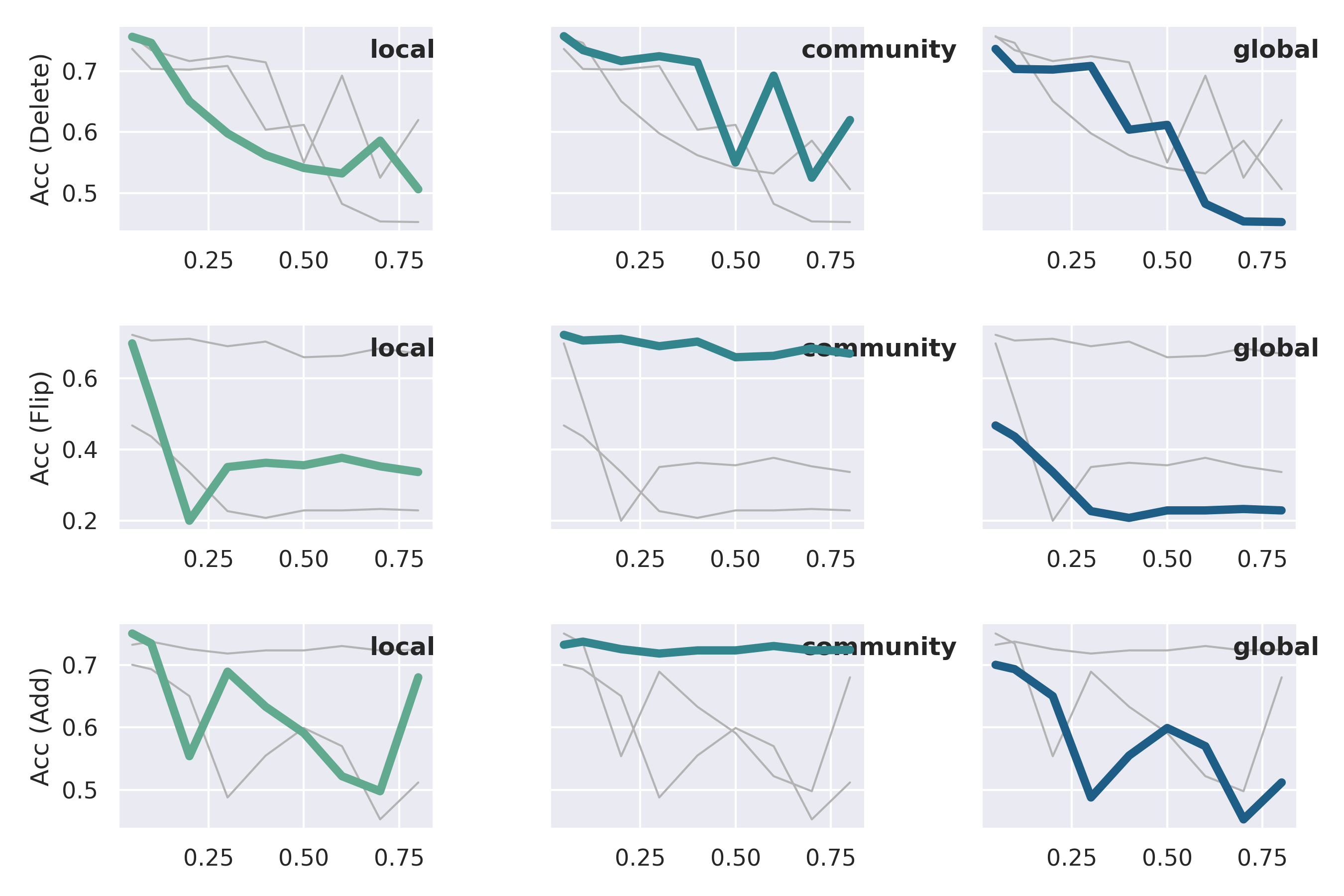}
  \caption{DropEdge on Citeseer.}
  \label{fig:dropedge_c}
  \vskip -0.2in
\end{figure}

\begin{figure}[htb]
  \centering
  \includegraphics[width=0.49\textwidth]{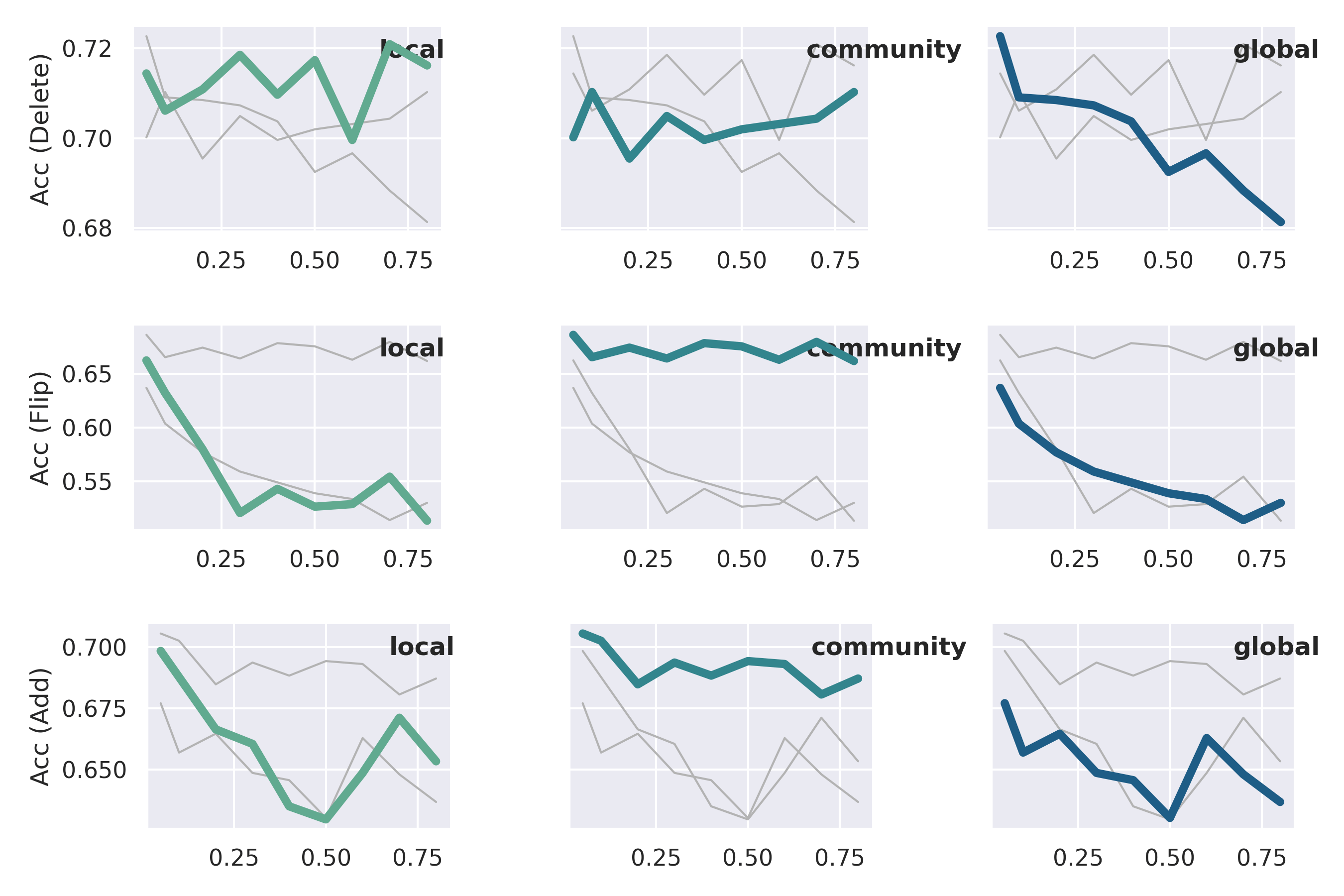}
  \caption{Pro-GNN on Citeseer.}
  \label{fig:prognn_c}
  \vskip -0.2in
\end{figure}

\begin{figure}[htb]
  \centering
  \includegraphics[width=0.49\textwidth]{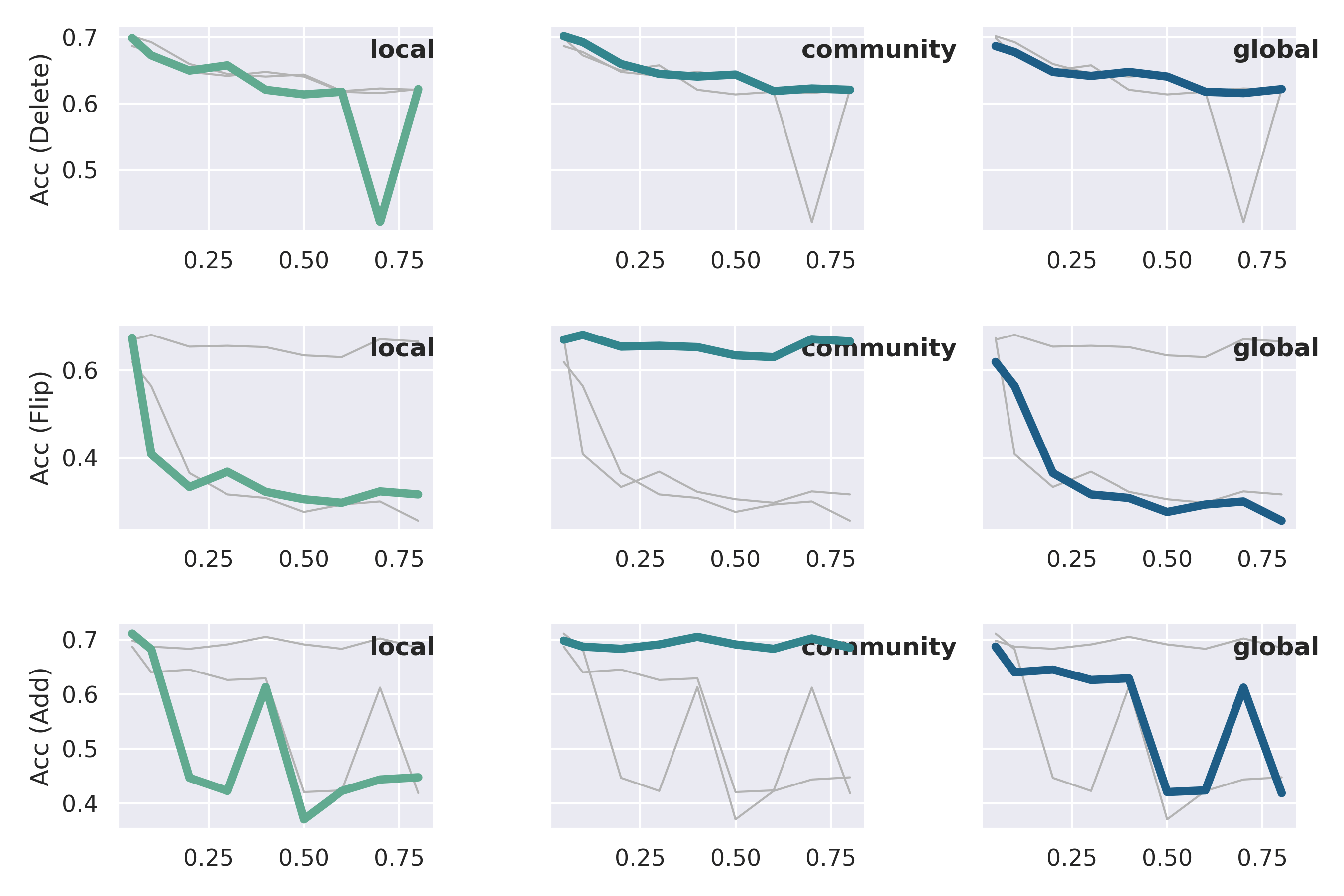}
  \caption{GCN on Citeseer.}
  \label{fig:gcn_c}
  \vskip -0.2in
\end{figure}

\begin{figure}[htb]
  \centering
  \includegraphics[width=0.49\textwidth]{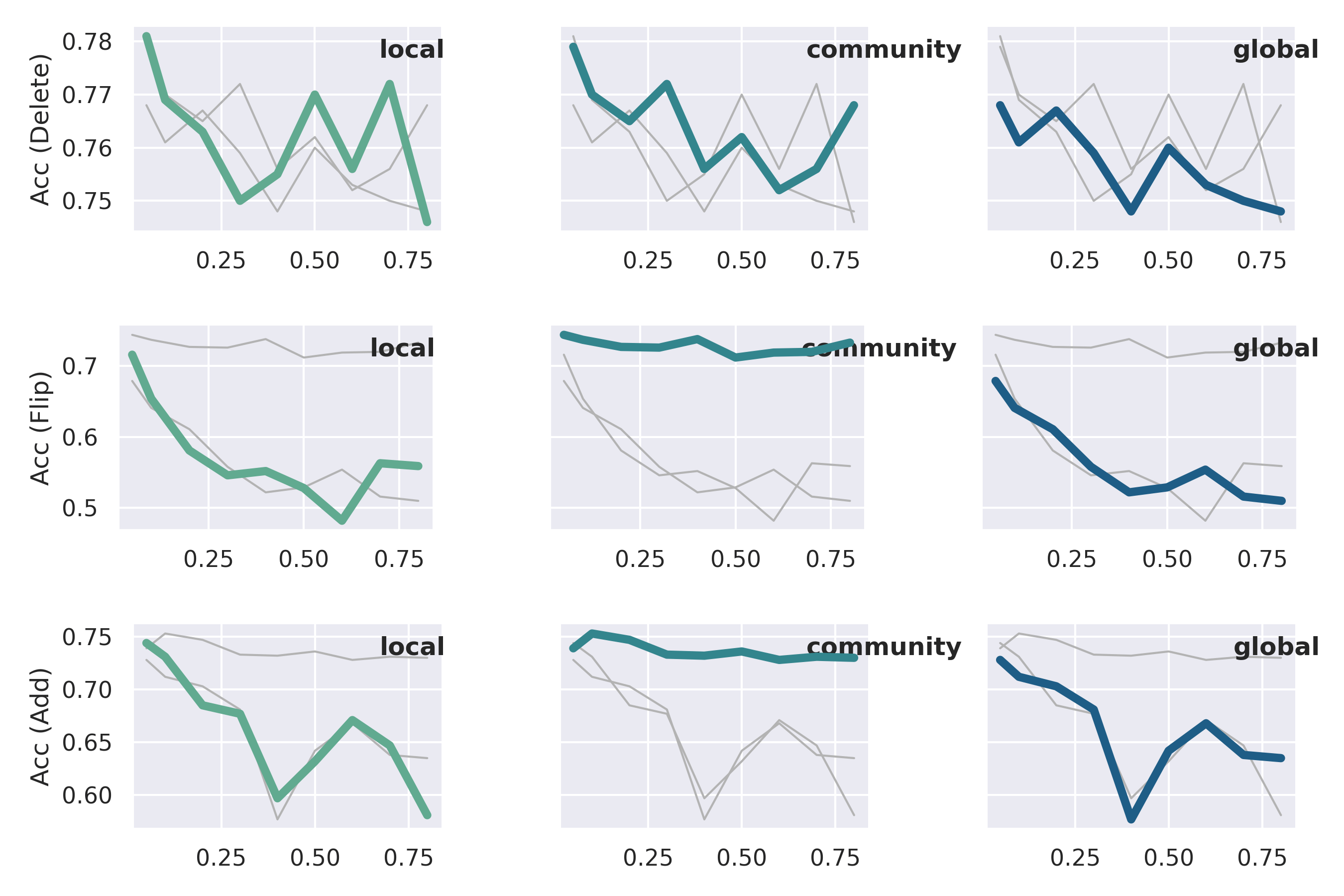}
  \caption{FastGCN on Citeseer.}
  \label{fig:fastgcn_c}
  \vskip -0.2in
\end{figure}

\begin{figure}[htb]
  \centering
  \includegraphics[width=0.49\textwidth]{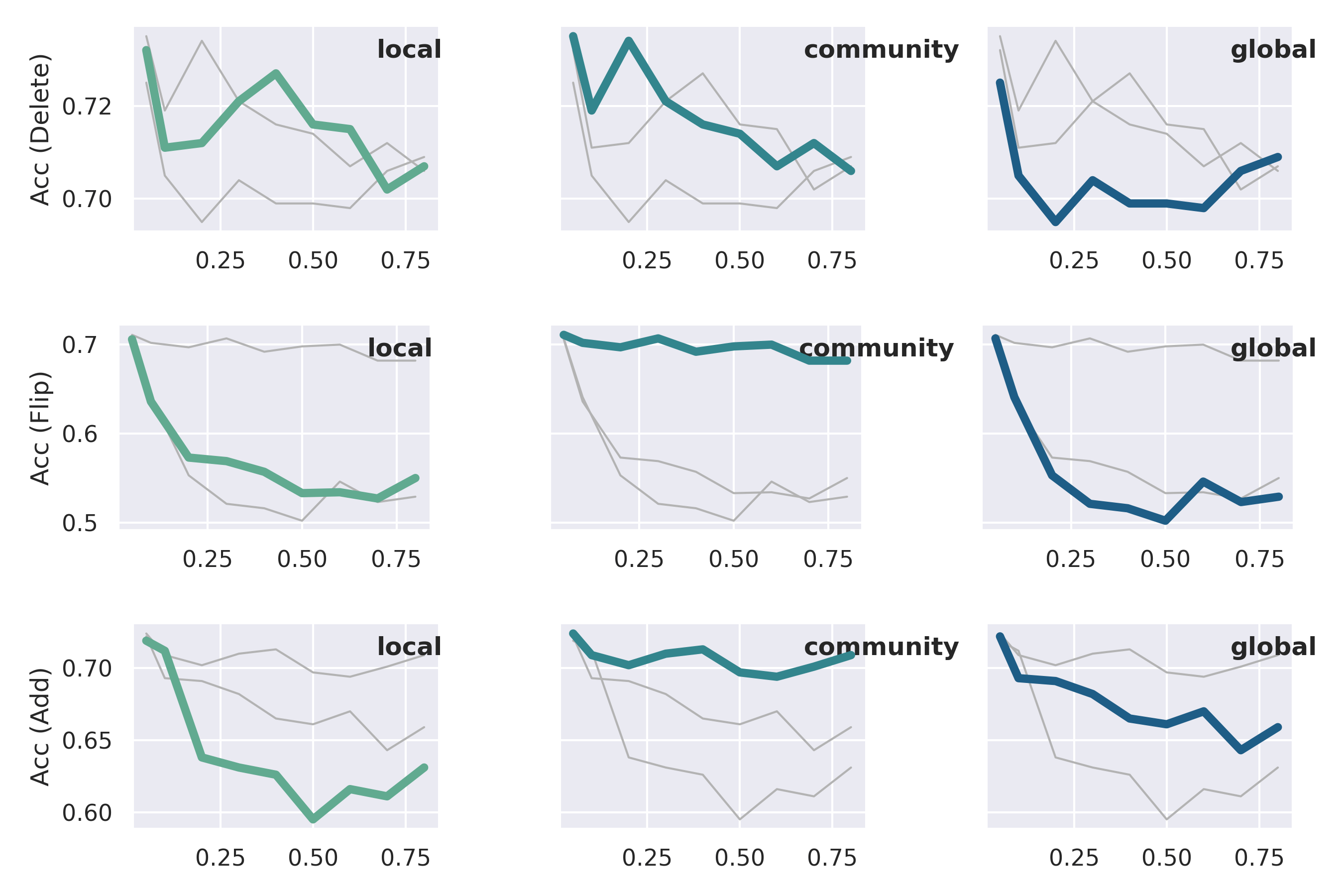}
  \caption{GRCN on Citeseer.}
  \label{fig:grcn_c}
  \vskip -0.2in
\end{figure}

\begin{figure}[htb]
  \centering
  \includegraphics[width=0.49\textwidth]{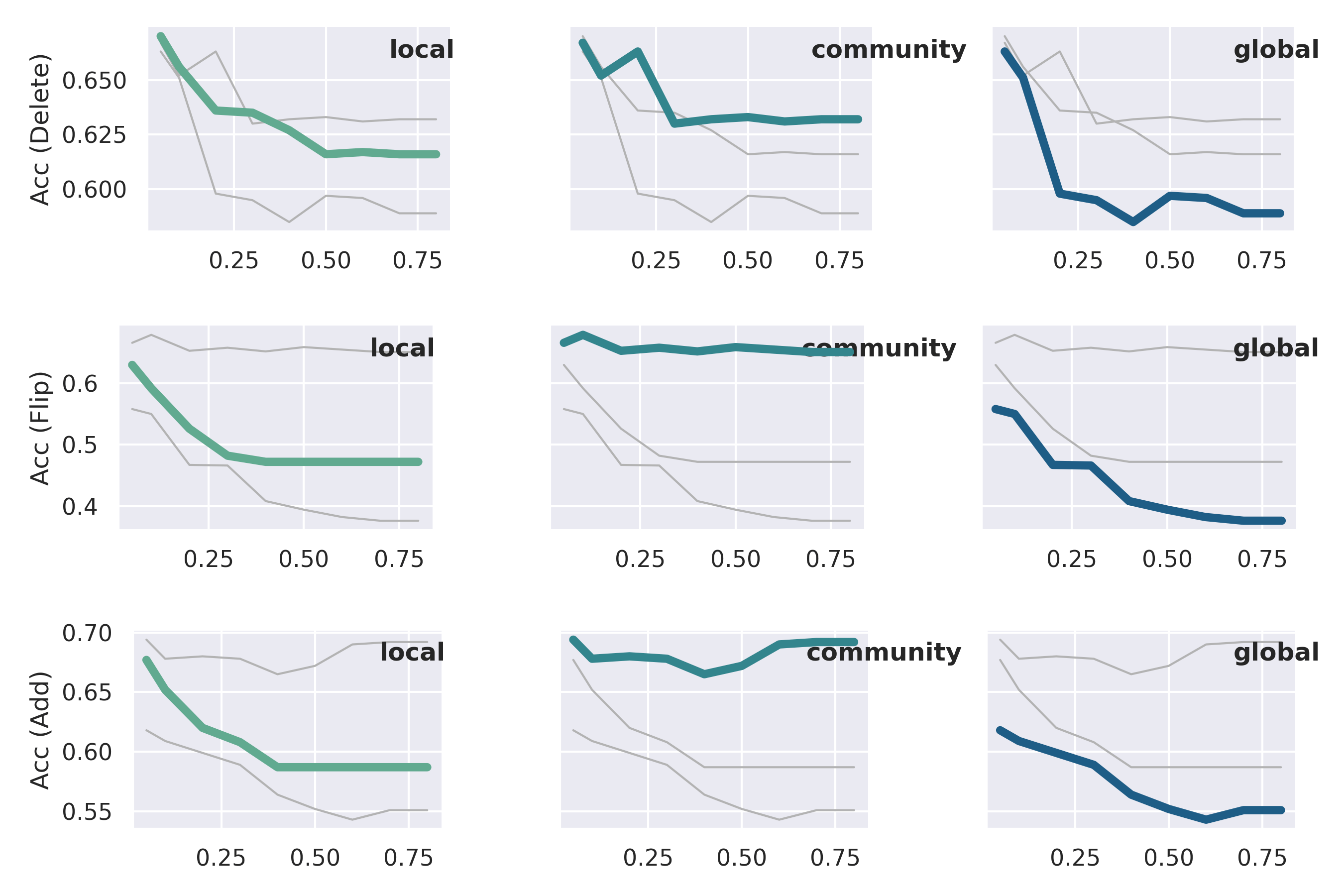}
  \caption{PTDNet on Citeseer.}
  \label{fig:ptdnet_c}
  \vskip -0.2in
\end{figure}

\begin{figure}[htb]
  \centering
  \includegraphics[width=0.49\textwidth]{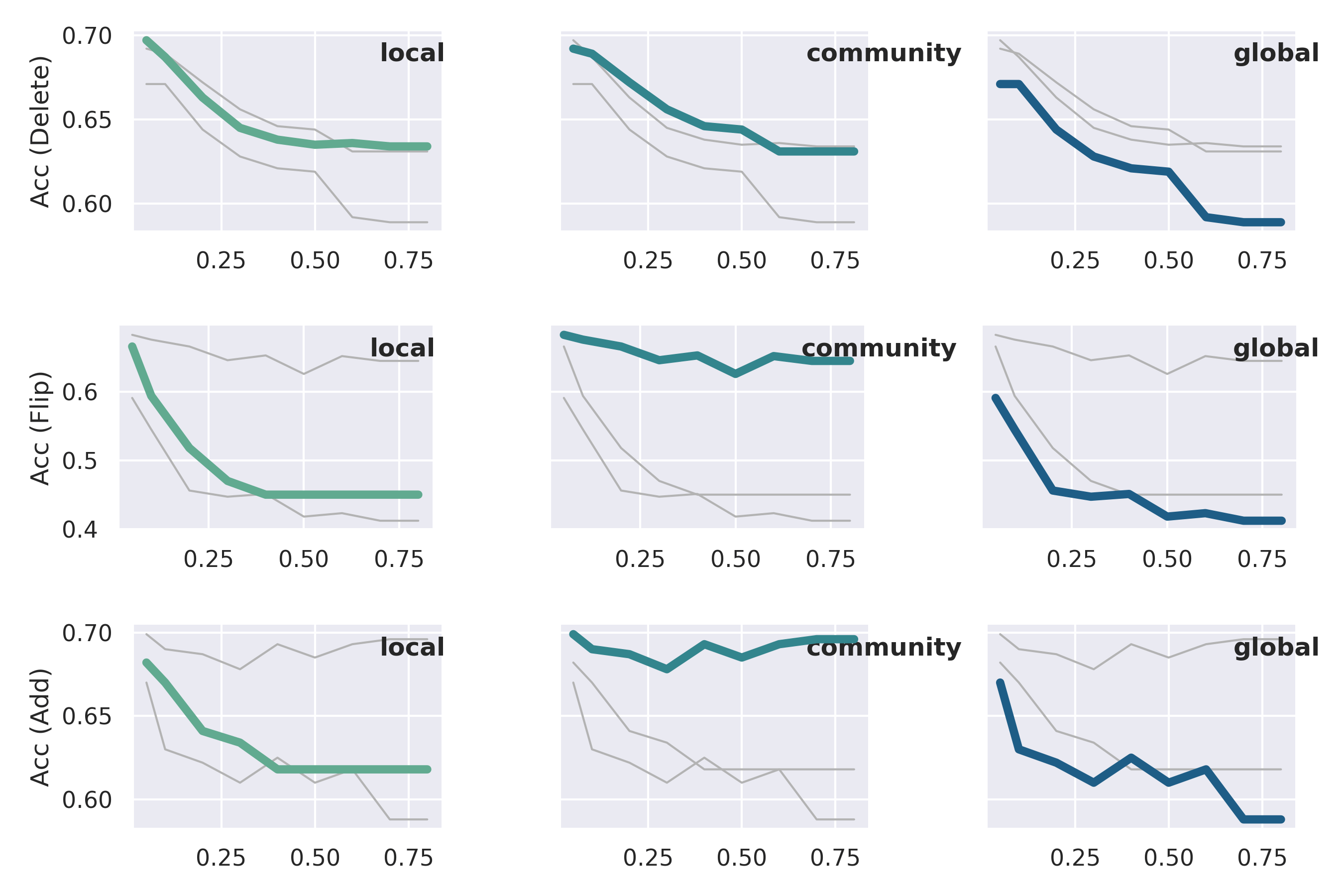}
  \caption{RobustGCN on Citeseer.}
  \label{fig:robustgcn_c}
  \vskip -0.2in
\end{figure}

The results of other robust GNNs are shown in Figure~\ref{fig:gcn} to~\ref{fig:robustgcn}. The overall classification accuracies of different GNNs to different structural noises on Citeseer are shown in Figure~\ref{overall-citeseer}. Similarly, we highlight every method in details in Figure~\ref{fig:graphsage_c} to~\ref{fig:robustgcn_c}.

\begin{table}[htb]
\vskip -0.1in
\centering
\caption{Brief statistics of used datasets.}
\label{dataset}
\begin{tabular}{c|c|c}
\hline
                                             & Cora                             & Citeseer                         \\\hline
number of labels                             & 7
    & 6                                             \\
number of nodes                              & 2708                             & 3312                              \\
number of edges                              & 5429                              & 4732                              \\
attribute dimension                          & 1433                              & 3703                              \\
isolated node                                & No                              & Yes                              \\
\multicolumn{1}{l|}{training/validation/test} & \multicolumn{1}{l|}{140/500/1000} & \multicolumn{1}{l}{120/500/1000}
\\\hline
\end{tabular}
\vskip -0.2in
\end{table}





\begin{figure*}[htb]
  \centering
  \includegraphics[width=1\textwidth]{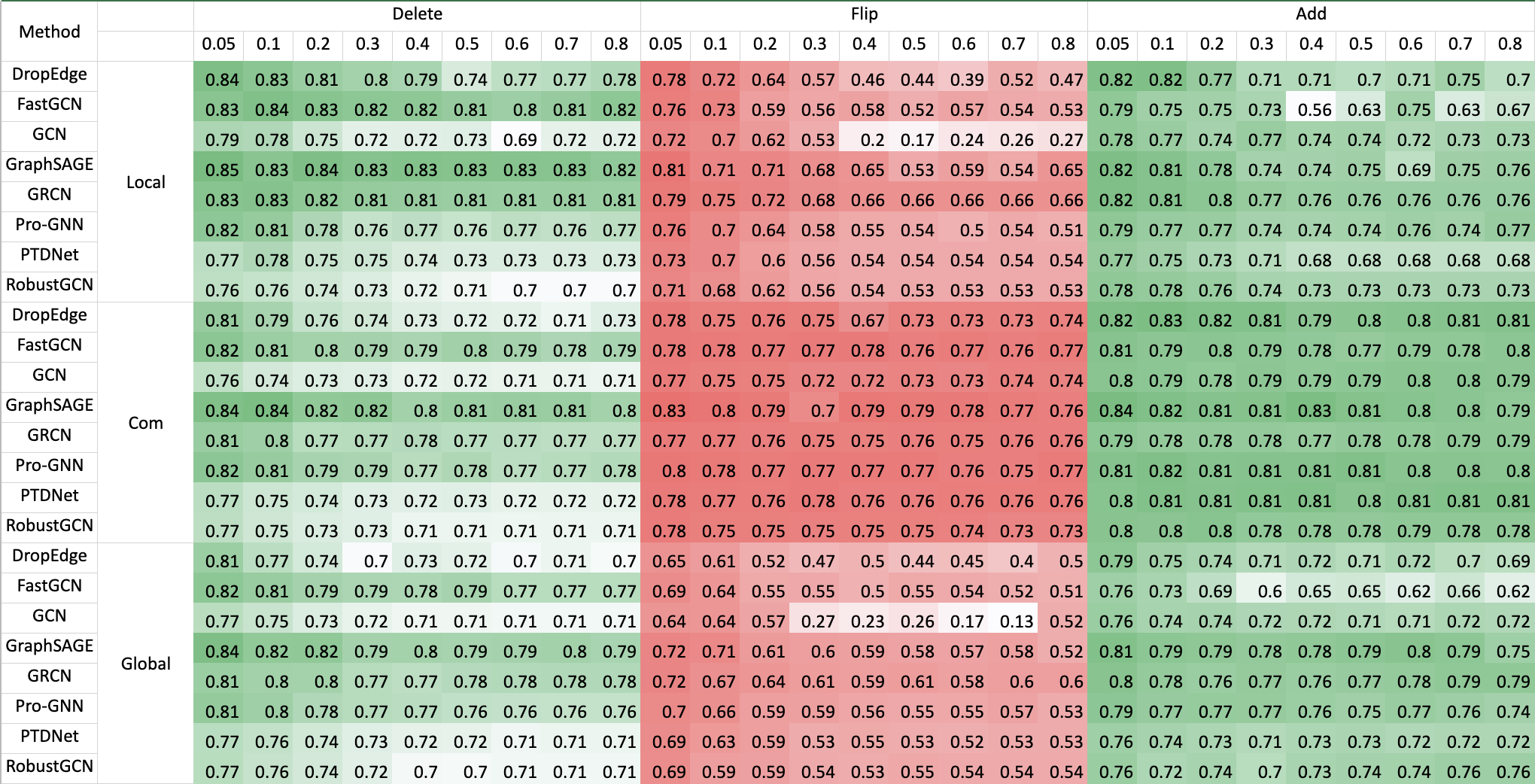}
  \caption{Part of experimental results on Cora.}
  \label{fig:full_rslt1}
  \vskip -0.2in
\end{figure*}

\begin{figure*}[htb]
  \centering
  \includegraphics[width=1\textwidth]{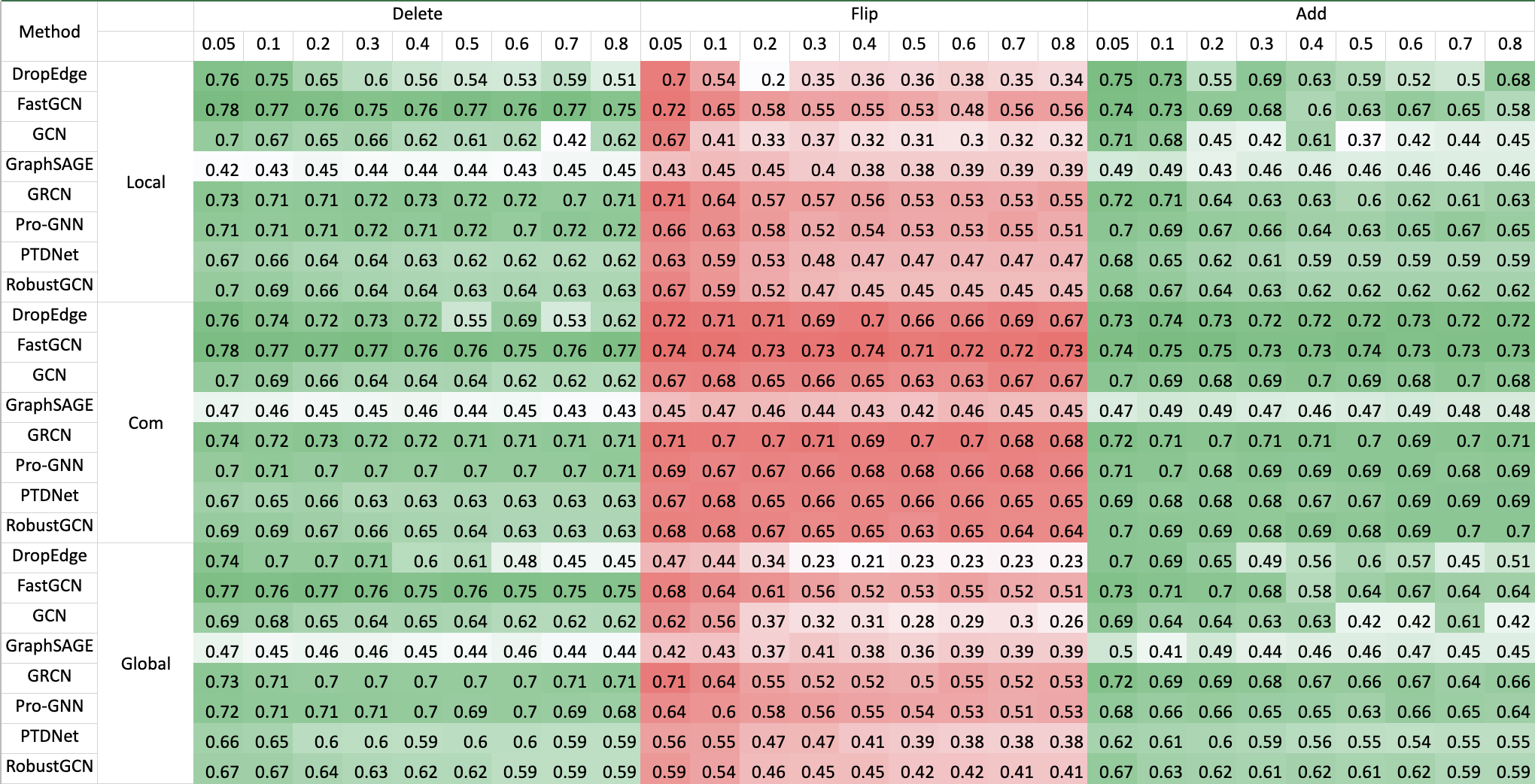}
  \caption{Part of experimental results on Citeseer.}
  \label{fig:full_rslt1}
  \vskip -0.2in
\end{figure*}

\end{document}